 \documentclass[pmlr,twocolumn,10pt]{jmlr} 

\usepackage{booktabs}
\usepackage{siunitx}

\usepackage{booktabs}
\usepackage{floatrow}
\usepackage{multicol}
\usepackage{multirow}
\usepackage[labelfont=bf]{caption}
\usepackage{adjustbox}
\usepackage{amsmath}
\usepackage{amssymb}
\usepackage{enumitem}
\usepackage{xcolor,colortbl}
\usepackage{bbding}
\usepackage[weather]{ifsym}
\usepackage[T1]{fontenc}
\usepackage[]{algorithm2e}
\usepackage[switch]{lineno} 

\DeclareMathOperator*{\argmin}{arg\,min}

\jmlrvolume{LEAVE UNSET}
\jmlryear{2023}
\jmlrsubmitted{LEAVE UNSET}
\jmlrpublished{LEAVE UNSET}
\jmlrworkshop{Machine Learning for Health (ML4H) 2023} 

\author{%
\Name{Keith Harrigian} \Email{kharrigian@jhu.edu}\\
\addr Johns Hopkins University, Department of Computer Science
\AND
\Name{Tina Tang} \Email{ttang19@jhu.edu}\\
\Name{Anthony Gonzales} \Email{agonzales@jhu.edu}\\
\Name{Cindy X. Cai} \Email{ccai6@jhmi.edu}\\
\addr Johns Hopkins School of Medicine, Wilmer Eye Institute
\AND
\Name{Mark Dredze} \Email{mdredze@cs.jhu.edu}\\
\addr Johns Hopkins University, Department of Computer Science
}

\title[An Eye on Clinical BERT]{An Eye on Clinical BERT: Investigating Language Model Generalization for Diabetic Eye Disease Phenotyping}

\begin{document}

\maketitle


\begin{abstract}
Diabetic eye disease is a major cause of blindness worldwide. The ability to monitor relevant clinical trajectories and detect lapses in care is critical to managing the disease and preventing blindness. Alas, much of the information necessary to support these goals is found only in the free text of the electronic medical record. To fill this information gap, we introduce a system for extracting evidence from clinical text of 19 clinical concepts related to diabetic eye disease and inferring relevant attributes for each. In developing this ophthalmology phenotyping system, we are also afforded a unique opportunity to evaluate the effectiveness of clinical language models at adapting to new clinical domains. Across multiple training paradigms, we find that BERT language models pretrained on out-of-distribution clinical data offer no significant improvement over BERT language models pretrained on non-clinical data for our domain. Our study tempers recent claims that language models pretrained on clinical data are necessary for clinical NLP tasks and highlights the importance of not treating clinical language data as a single homogeneous domain.
\end{abstract}



\section{Introduction} \label{sec:intro}

Diabetic eye disease (e.g., diabetic retinopathy, diabetic macular edema) is a major cause of blindness worldwide \citep{steinmetz2021causes,wykoff2021risk}. These conditions can develop in patients with diabetes, whereby elevated sugar levels in the blood can cause damage to the retinal blood vessels. Management of diabetic eye disease relies not only on a patient’s glycemic control, but also on regular ophthalmic screening for the early detection and treatment of vision threatening complications \citep{solomon2017diabetic,flaxel2020diabetic}. The ability to monitor clinical trajectories and efficiently detect lapses in care is critical to achieving the latter objectives. Unfortunately, structured data in the electronic health record (EHR) remains ill-suited for describing many ophthalmic conditions at the granularity necessary to support these goals \citep{cai2021effect}. Much of the critical information is found only in the free text of the EHR.

Provided sufficient support at training time, supervised machine learning models can extract useful clinical information from free text in the EHR to augment structured metadata \citep{voorham2007computerized,mccoy2017enhancing,koleck2019natural}. However, annotated clinical datasets are typically small by contemporary standards due to the inherent bottleneck imposed by the necessity of involving highly trained domain experts (i.e., physicians and other healthcare professionals) \citep{spasic2020clinical}. This limitation is compounded in the diabetic eye disease use case, where the number of clinical concepts needed to effectively monitor the condition is large \citep{pearce2019association,gale2021diabetic} and their associated attributes (e.g., severity, temporality) are heavily class imbalanced \citep{yau2012global,yang2019prevalence}. 

Language models (LMs) pretrained on massive text corpora are a powerful tool for representing language across a variety of downstream modeling tasks \citep{howard2018universal,wei2021pretrained}. In low-resource settings particularly, pretraining can inject useful knowledge for differentiating linguistic instances in context \citep{gao2021making}. This fact has inspired the training and release of several models trained on biomedical and clinical text over the last several years \citep{lee2020biobert,gu2021domain}. 

Nonetheless, the success of LMs as encoders in the clinical domain and beyond is typically correlated with the degree of alignment between pretraining and task-specific language distributions \citep{roberts2016assessing,gururangan2020don,talmor2020olmpics}. The field of ophthalmology serves as a departure from the domains on which existing clinical language models have been trained \citep{alsentzer2019publicly,yang2022gatortron}, requiring highly detailed knowledge of a single anatomical system. In developing a phenotyping system for diabetic eye disease, we find ourselves in a unique position to answer an important question: Do clinical LMs perform consistently better than non-clinical LMs on out-of-domain clinical data?




In the remainder of this paper, we conduct an empirical investigation of multiple BERT encoders and training paradigms, allowing us to evaluate the sufficiency of existing BERT language models in a specialized clinical domain. In contrast to common perceptions about clinical language models, we find that the LMs trained on out-of-domain clinical data provide little-to-no benefit in our domain compared to the LMs trained on non-clinical data. Furthermore, advantages derived from an initial pretraining phase can be nullified almost entirely via tailored in-domain pretraining. Given the ubiquity of distribution shift and scarcity of data across clinical NLP use cases, our results suggest that the research community may benefit from focusing on adapting language models to low-resource clinical settings instead of training ``general'' clinical language models from scratch.



\section{Background} \label{sec:related}

\subsection{Diabetic Eye Disease}

Diabetic eye disease refers to a collection of eye problems that can result from diabetes, including diabetic retinopathy (DR) and diabetic macular edema (DME) \citep{solomon2017diabetic,flaxel2020diabetic}. DR is a progressive disease caused by insufficient blood flow to the retina which, in its most severe state, sees the growth of abnormal blood vessels around the retina. This process, referred to generally as neovascularization, may lead to vision-threatening complications such as vitreous hemorrhage, retinal detachment, and blindness \citep{steinmetz2021causes}. 

Although advanced stages of diabetic eye disease cannot be reversed, treatments can prevent the condition from worsening and even return some visual fidelity if delivered in a timely manner \citep{duh2017diabetic}. Follow-up timelines depend on patient-specific trajectories which are specified only with the free text of the EHR \citep{cai2021effect}. Extraction and synthesis of this information has the potential to dramatically reduce the rate at which patients are lost to follow up, for example by introducing automatic notifications regarding delayed treatment \citep{gale2021diabetic}.

\subsection{NLP in Ophthalmology} \label{sec:related-oph}

There exists a brief, albeit rich, history of artificial intelligence systems targeting problems in the field of ophthalmology \citep{grewal2018deep,ting2019artificial}. The majority of effort has been allocated to improving imaging diagnostics via computer vision \citep{teikari2019embedded} and building clinical decision support systems using structured EHR data \citep{ogunyemi2021detecting,jacoba2021biomarkers}. Systems leveraging natural language processing (NLP) techniques to improve ophthalmic care make up the minority of these efforts and have typically focused on narrow concept extraction objectives \citep{liu2017natural,mao2017named}. However, the rise of large LMs, such as GPT-3, has drawn increased attention to the NLP research community from ophthalmologists interested in better synthesizing free text in the EHR \citep{yang2021deep,nath2022new}. 


The most similar work to our own comes from \citet{yu2022identify}. Although our work shares a common objective in extracting concepts related to diabetic retinopathy and linking their associated attributes, there are multiple key differences. First, \citet{yu2022identify} focus on imaging reports from patients already diagnosed with diabetic retinopathy, whereas we focus on progress notes and problem lists from a general ophthalmology patient population. Second, our ontology of clinical concepts is larger and more diverse (e.g., we include comorbidities and treatments). Finally, they approach the assignment of attributes to clinical concepts as a relation extraction task, which assumes overt evidence of each attribute in the free text. In contrast, we assume some attributes are not explicitly stated in the text, but can be inferred based on context and reasoning.


\subsection{Clinical Language Modeling} \label{sec:related-lm}

Our understanding of the value of clinical language models has evolved significantly over time, but remains far from complete.  Early work in neural language modeling demonstrated that word embeddings learned using clinical and biomedical text can improve performance in downstream clinical tasks compared to embeddings trained on general web data \citep{wu2015clinical,dingwall2018mittens}. Similar results have emerged for contextual language models \citep{khattak2019survey,alsentzer2019publicly,lee2020biobert}.

Why do these clinical language models typically outperform their generic counterparts on clinical tasks? The primary hypothesis is that pretraining on clinical data is necessary to address the distributional shift that occurs from non-clinical to clinical settings \citep{naik2021adapting,lamproudis2022evaluating}. Common examples include changes in the distribution of word senses (e.g., aggressive treatment regimen, aggressive behavior) and the introduction of medicine-specific terminology (e.g., abbreviations, diagnoses, etc.) \citep{wu2011semantic,liu2012towards}. Research from \citet{lewis2020pretrained} and \citet{lehman2023we} has suggested that domain-specific vocabularies are invaluable for allowing clinical language models to learn semantics in a more parameter-efficient manner. Recently, these tenets have motivated the development of domain-specific GPT-style models outside of the clinical space \citep{taylor2022galactica,venigalla2022biomedlm,wu2023bloomberggpt}.

At the same time, other researchers have shown that non-clinical language models, provided sufficient size, are still able to perform remarkably well in clinical and clinical-adjacent tasks \citep{agrawal2022large,singhal2022large,harrigian2023characterization}. Moreover, domain-specific vocabularies tailored for clinical tasks do not consistently provide comprehensive performance improvements \citep{gutierrez2023biomedical}. Are these negative results an anomaly? Or do they more accurately represent the capabilities of clinical language models?

Unfortunately, challenges with sharing sensitive clinical data have thus far limited the strength of conclusions we can draw regarding the value of clinical language models. Many of the public datasets that are used to evaluate clinical language models are drawn from the same public datasets used to train the language models and generally cover a narrow range of clinical tasks \citep{thirunavukarasu2023large,wornow2023shaky}. It is unclear whether available clinical language models are better on clinical data broadly, or on the specific medical speciality for which they are trained.

In this study, we leverage a unique clinical dataset to evaluate the sufficiency of available clinical language models on a new clinical domain. Since not all clinical datasets are drawn from the same domain, it is important to determine the advantages available clinical models provide for work on the diverse range of clinical domains.

\section{Data} \label{sec:data}

To the best of our knowledge, only one EHR dataset containing diabetic eye disease annotations exists \citep{yu2022identify}. In addition to not being publicly available, this dataset has several shortcomings that make it suboptimal for our use case (e.g., biased patient population, imaging report focus, extractive modeling setting, \S \ref{sec:related-oph}). To ensure that we can cover the breadth of concepts and attributes required to monitor diabetic eye disease in our patient population, we curate a new clinical note dataset from scratch. Below, we discuss the design hurdles experienced throughout the dataset development process and subsequent decisions undertaken to address them.

\subsection{Inclusion Criteria} \label{sec:data-inclusion}

All ophthalmology-related visits to our institute's hospital system from January 1, 2013 through April 1, 2022 were considered candidates for the study. Visits for imaging services and visits which lacked either a progress note or problem list \citep{weed1968medical} were excluded, leaving a total of 692,486 visits by 91,097 patients. Notes were processed adhering to our institution's privacy policy after approval by our Institutional Review Board (IRB).

\subsection{Concept Ontology} \label{sec:data-ontology}

We developed a multi-level ontology for 19 clinical concepts with significance in the management and treatment of diabetic eye disease. Concepts with similar clinical relevance were grouped together into higher-level semantic categories (e.g., Retina Conditions, Complications of Diabetes Mellitus). Each concept was further associated with modifiers within up to 3 attribute categories (i.e., Laterality, Temporality, and Severity/Type). The full ontology is provided in Table \ref{tab:ontology} of Appendix \ref{apx:data-ontology}.

\begin{figure*}[t]
    \centering\resizebox{\textwidth}{!}{%
    \begin{tabular}{llllllllll} \toprule
        \multicolumn{10}{l}{\textbf{Document ID}: ad1fc53fe509fdea65d2099d8b5b3c57a8b5d1978f9bb8f0fa5eb1c427015aaf} \\
        \multicolumn{10}{l}{\textbf{Encounter Date}: 2015-06-21} \\ \toprule
        \multicolumn{10}{l}{} \\
        \multicolumn{10}{l}{[[[ENCOUNTER ICD-10 CODES]]]} \\
        \multicolumn{10}{l}{[[E11.319: Diabetic retinopathy]]} \\
        \multicolumn{10}{l}{[[E11.311: Diabetic macular edema, both eyes]]} \\
        \multicolumn{10}{l}{} \\
        \multicolumn{10}{l}{[[[PROBLEM LIST]]]} \\
        \multicolumn{10}{l}{[[E11.3313: Diabetic macular edema of both eyes with moderate nonproliferative diabetic retinopathy associated with Type 2 diabetes mellitus]]} \\
        \multicolumn{10}{l}{[OVERVIEW]} \\
        \multicolumn{10}{l}{Eylea initiated right eye 6/2015 and and left eye 5/2014. No progression to PDR.} \\
        \multicolumn{10}{l}{[ASSESSMENT \& PLAN]} \\
        \multicolumn{10}{l}{Right eye has foveal edema. Eylea \#1 given. She will return in 2 weeks for eylea left eye after vacation to MI.} \\
        \multicolumn{10}{l}{} \\ \toprule
        \textbf{Start} & \textbf{End} & \textbf{Concept} & \textbf{Text Span} & \textbf{Context} & \textbf{Laterality} & \textbf{Severity/Type} & \textbf{Temporality} & \textbf{Negated} & \textbf{Incorrect} \\ \toprule
        31 & 38 & DR (General) & E11.319 & [[<<E11.319>>: Diabetic Retinopathy]] & $^\blacktriangledown$ OU & \multicolumn{1}{c}{--} & $^\blacktriangledown$ Active & $^\blacktriangledown$ & $^\blacktriangledown$ \\ \midrule
        31 & 38 & DM & E11.319 & [[<<E11.319>>: Diabetic Retinopathy]] & \multicolumn{1}{c}{--} & $^\blacktriangledown$ Type 2 & $^\blacktriangledown$ Active & $^\blacktriangledown$ & $^\blacktriangledown$ \\ \midrule
        267 & 270 & PDR & PDR & left eye 5/2014. No progression to <<PDR>>. & $^\blacktriangledown$ OU & $^\blacktriangledown$  & $^\blacktriangledown$ Active & $^\blacktriangledown$ Negated & $^\blacktriangledown$ \\ \midrule
        \multicolumn{10}{c}{$\cdots$} \\ \midrule
        525 & 537 & ME & foveal edema & Right eye has <<foveal edema>>. Eylea \#1 & $^\blacktriangledown$ OD & $^\blacktriangledown$ CI-DME & $^\blacktriangledown$ Active & $^\blacktriangledown$ & $^\blacktriangledown$ \\ \midrule
        601 & 603 & Heart Attack & MI & eylea left eye after vacation to <<MI>>. & \multicolumn{1}{c}{--} & \multicolumn{1}{c}{--} & $^\blacktriangledown$ & $^\blacktriangledown$ & $^\blacktriangledown$ Incorrect \\ \bottomrule
    \end{tabular}}
    \caption{Interface displayed to annotators in Microsoft Excel. Drop-down data validation cells provide possible attribute labels conditioned on each row's clinical concept, with irrelevant attributes denoted using a `--' symbol. If an attribute is not clearly specified within a note or not inferable via context (e.g., severity of PDR), the annotator is instructed to leave the cell blank; we treat these instances as missing data at training time.}
    \label{fig:annotationfile}
\end{figure*}

\subsection{Annotation} \label{sec:data-annotation}

Ophthalmology notes commonly refer to the same clinical concept multiple times, albeit with different attributes, thus rendering note-level application of our ontology inappropriate. Span-level annotation was necessary, but non-trivial. Within pilot experiments, our domain experts found it challenging to consistently identify spans across the relatively wide label space. Moreover, data privacy and technical limitations made it infeasible to deploy existing annotation software capable of both span identification and labeling.

As an alternative strategy, we curated a set of high-recall regular expressions to identify candidate concept spans which could then be shown to annotators to validate correctness and assign appropriate attribute labels. Expressions were applied to the free text of the note first, and then to diagnostic codes (i.e., International Classification of Diseases, Tenth Revision (ICD-10)) contained in the problem list and note metadata. To limit redundant annotation efforts, concepts found in the diagnostic codes were excluded if already found in the free text. Patterns for the free-text were developed iteratively with domain experts, while relevant diagnostic codes were identified using an online database.\footnote{\url{https://www.icd10data.com}}



Concept matches were organized by encounter and displayed in context to facilitate span-level attribute annotation (see Figure \ref{fig:annotationfile}). Two domain experts (a post-graduate year-4 ophthalmology resident and a licensed optometrist) independently reviewed notes for 736 clinical encounters from a random sample of 348 patients.\footnote{Institutional policy limited the number of patients whose notes could be accessed at a time for annotation.} Disagreements were resolved through discussion, with oversight from a board-certified ophthalmologist. A total of 12,723 attribute labels were generated from 6,565 spans (see Table \ref{tab:labeldist} in Appendix \ref{apx:data-extract}). Additional label statistics and an analysis of our concept extraction protocol's sensitivity can be found in Appendix \ref{apx:data-extract}. All annotation was completed using programatically-generated Microsoft Excel workbooks which could be deployed in a HIPAA-compliant remote desktop environment. Although we are unable to release our annotations due to privacy constraints, we have made available our set of regular expressions and code used for generating the annotation workbooks.\footnote{\url{https://github.com/kharrigian/ml4h-clinical-bert}}


\subsection{Task Consolidation} \label{sec:data-tasks}

The resolved set of annotations exhibits significant class and concept imbalance -- some attribute classes from the ontology are not even represented in the dataset. To ensure our machine learning system is able to effectively extract signal during training, we consolidate certain attribute classes and group together semantically similar concept-attribute pairs. The remapped ontology breaks down into 14 unique classification tasks. The mapping was constructed with our downstream clinical use case in mind, ensuring to preserve the minimum resolution necessary to monitor trajectories related to diabetic eye disease. The mapping from ontology to task space is included in Table \ref{tab:mapping} of Appendix \ref{apx:data-tasks}.


\begin{table*}[t]
    \centering
    \setlength{\tabcolsep}{2.5pt}
    \begin{adjustbox}{width=\linewidth}
    \begin{tabular}{ c l c c c c c c c c c} \toprule
        & 
            & 
            & 
            \multicolumn{4}{c}{\textbf{BERT Base}} &
            \multicolumn{4}{c}{\textbf{Clinical BERT}} \\ \cmidrule(lr){4-7} \cmidrule(lr){8-11}
        & 
            & 
            &
            \multicolumn{2}{c}{\begin{tabular}{@{}c@{}}\textbf{w/o Continued}\\\textbf{Pretraining}\end{tabular}} & 
            \multicolumn{2}{c}{\begin{tabular}{@{}c@{}}\textbf{w/ Continued}\\\textbf{Pretraining}\end{tabular}} &
            \multicolumn{2}{c}{\begin{tabular}{@{}c@{}}\textbf{w/o Continued}\\\textbf{Pretraining}\end{tabular}} &
            \multicolumn{2}{c}{\begin{tabular}{@{}c@{}}\textbf{w/ Continued}\\\textbf{Pretraining}\end{tabular}}
             \\ \cmidrule(lr){4-5} \cmidrule(lr){6-7} \cmidrule(lr){8-9} \cmidrule(lr){10-11}
         \textbf{Attribute} & 
            \multicolumn{1}{l}{\textbf{Concepts (\# Classes)}} &  
            \multicolumn{1}{c}{\textbf{Majority}} & 
            \raisebox{-0.5ex}{\SnowflakeChevron} & 
            \raisebox{-0.5ex}{\FilledRainCloud} & 
            \raisebox{-0.5ex}{\SnowflakeChevron} & 
            \raisebox{-0.5ex}{\FilledRainCloud} & 
            \raisebox{-0.5ex}{\SnowflakeChevron} & 
            \raisebox{-0.5ex}{\FilledRainCloud} & 
            \raisebox{-0.5ex}{\SnowflakeChevron} & 
            \raisebox{-0.5ex}{\FilledRainCloud} \\ \midrule
         \multirow{3}{*}{Temporality} & 
            Retina ($K$=2) & 
            .76 \textsubscript{(.75,.77)} &
            .81 \textsubscript{(.79,.82)} &
            .83 \textsubscript{(.82,.84)} &
            .84 \textsubscript{(.83,.86)} &
            .87 \textsubscript{(.85,.89)} &
            .83 \textsubscript{(.82,.84)} &
            .84 \textsubscript{(.83,.86)} &
            .85 \textsubscript{(.84,.87)} &
            .87 \textsubscript{(.85,.88)} \\ 
         & 
            DM Complications ($K$=2) & 
            .81 \textsubscript{(.72,.87)} &
            .81 \textsubscript{(.73,.88)} &
            .80 \textsubscript{(.70,.89)} &
            .81 \textsubscript{(.71,.88)} &
            .80 \textsubscript{(.71,.88)} &
            .80 \textsubscript{(.70,.89)} &
            .84 \textsubscript{(.76,.90)} &
            .84 \textsubscript{(.73,.93)} &
            .85 \textsubscript{(.77,.92)} \\
         & 
            Treatment ($K$=3) & 
            .35 \textsubscript{(.34,.36)} &
            .59 \textsubscript{(.55,.64)} &
            .79 \textsubscript{(.75,.82)} &
            .81 \textsubscript{(.79,.83)} &
            .84 \textsubscript{(.81,.86)} &
            .69 \textsubscript{(.65,.73)} &
            .81 \textsubscript{(.78,.84)} &
            .81 \textsubscript{(.77,.83)} &
            .82 \textsubscript{(.76,.85)} \\ \cmidrule{1-11}
         \multirow{1}{*}{Laterality} & 
            All ($K$=3) & 
            .53 \textsubscript{(.52,.55)} &
            .54 \textsubscript{(.51,.57)} &
            .84 \textsubscript{(.83,.86)} &
            .60 \textsubscript{(.58,.61)} &
            .92 \textsubscript{(.90,.93)} &
            .56 \textsubscript{(.54,.58)} &
            .84 \textsubscript{(.81,.87)} &
            .60 \textsubscript{(.57,.63)} &
            .90 \textsubscript{(.89,.92)} \\ \cmidrule{1-11}
         \multirow{6}{*}{Type} & 
            ME ($K$=2) & 
            .83 \textsubscript{(.77,.88)} &
            .83 \textsubscript{(.79,.88)} &
            .87 \textsubscript{(.79,.93)} &
            .86 \textsubscript{(.79,.91)} &
            .88 \textsubscript{(.82,.94)} &
            .82 \textsubscript{(.78,.86)} &
            .87 \textsubscript{(.80,.93)} &
            .85 \textsubscript{(.82,.89)} &
            .90 \textsubscript{(.85,.94)} \\
         & RD ($K$=4) & 
            .68 \textsubscript{(.52,.85)} &
            .75 \textsubscript{(.57,.89)} &
            .79 \textsubscript{(.58,.95)} &
            .70 \textsubscript{(.55,.83)} &
            .82 \textsubscript{(.76,.88)} &
            .78 \textsubscript{(.59,.94)} &
            .81 \textsubscript{(.61,.98)} &
            .72 \textsubscript{(.50,.95)} &
            .87 \textsubscript{(.78,.95)} \\
         & 
            NV ($K$=5) & 
            .79 \textsubscript{(.63,.91)} &
            .66 \textsubscript{(.53,.80)} &
            .75 \textsubscript{(.60,.89)} &
            .81 \textsubscript{(.71,.91)} &
            .82 \textsubscript{(.72,.93)} &
            .71 \textsubscript{(.56,.87)} &
            .78 \textsubscript{(.65,.92)} &
            .81 \textsubscript{(.73,.90)} &
            .77 \textsubscript{(.66,.86)}\\
         & 
            DM ($K$=3) & 
            .41 \textsubscript{(.30,.55)} &
            .39 \textsubscript{(.31,.53)} &
            .40 \textsubscript{(.32,.52)} &
            .37 \textsubscript{(.29,.52)} &
            .57 \textsubscript{(.43,.79)} &
            .31 \textsubscript{(.28,.35)} &
            .40 \textsubscript{(.31,.53)} &
            .33 \textsubscript{(.29,.38)} &
            .54 \textsubscript{(.40,.74)} \\
         & 
            NVG Surgery ($K$=3) &
            .91 \textsubscript{(.76,1.0)} &
            .85 \textsubscript{(.69,1.0)} &
            .85 \textsubscript{(.69,1.0)} &
            .74 \textsubscript{(.63,.88)} &
            .85 \textsubscript{(.69,1.0)} &
            .79 \textsubscript{(.54,1.0)} &
            .85 \textsubscript{(.69,1.0)} &
            .79 \textsubscript{(.63,.96)} &
            .85 \textsubscript{(.69,1.0)} \\
         & 
            Retina Surgery ($K$=2) & 
            .58 \textsubscript{(.47,.68)} &
            .51 \textsubscript{(.38,.64)} &
            .66 \textsubscript{(.52,.79)} &
            .60 \textsubscript{(.44,.71)} &
            .71 \textsubscript{(.59,.83)} &
            .59 \textsubscript{(.46,.70)} &
            .52 \textsubscript{(.46,.59)} &
            .64 \textsubscript{(.53,.75)} &
            .76 \textsubscript{(.65,.85)} \\ \cmidrule{1-11}
         \multirow{2}{*}{Severity} & 
            NPDR ($K$=3)  & 
            .54 \textsubscript{(.48,.61)} &
            .56 \textsubscript{(.50,.62)} &
            .83 \textsubscript{(.73,.90)} &
            .69 \textsubscript{(.57,.84)} &
            .89 \textsubscript{(.77,.98)} &
            .58 \textsubscript{(.54,.61)} &
            .91 \textsubscript{(.87,.96)} &
            .70 \textsubscript{(.60,.83)} &
            .95 \textsubscript{(.90,.99)} \\
         & 
            PDR ($K$=2)  & 
            .48 \textsubscript{(.47,.49)} &
            .45 \textsubscript{(.43,.48)} &
            .71 \textsubscript{(.53,.89)} &
            .39 \textsubscript{(.29,.47)} &
            .81 \textsubscript{(.64,.93)} &
            .43 \textsubscript{(.37,.47)} &
            .53 \textsubscript{(.36,.77)} &
            .45 \textsubscript{(.43,.48)} &
            .82 \textsubscript{(.63,.98)} \\ \cmidrule{1-11}
         \multirow{2}{*}{Span Validity} & 
            ME ($K$=2) &
            .60 \textsubscript{(.49,.80)} &
            .60 \textsubscript{(.49,.72)} &
            .55 \textsubscript{(.49,.62)} &
            .77 \textsubscript{(.57,.97)} &
            .81 \textsubscript{(.64,.97)} &
            .65 \textsubscript{(.52,.78)} &
            .56 \textsubscript{(.50,.64)} &
            .66 \textsubscript{(.49,.86)} &
            .83 \textsubscript{(.65,.98)} \\
         & 
            Retina Surgery ($K$=2) &
            .46 \textsubscript{(.44,.47)} &
            .77 \textsubscript{(.72,.81)} &
            .77 \textsubscript{(.68,.87)} &
            .83 \textsubscript{(.77,.89)} &
            .82 \textsubscript{(.75,.90)} &
            .76 \textsubscript{(.69,.84)} &
            .80 \textsubscript{(.74,.86)} &
            .83 \textsubscript{(.76,.90)} &
            .80 \textsubscript{(.75,.88)} \\ \cmidrule{1-11}
        \multicolumn{2}{c}{\textbf{Average (All Tasks)}} &
            .62 \textsubscript{(.58,.67)} &
            .65 \textsubscript{(.61,.69)} &
            .75 \textsubscript{(.70,.79)} &
            .70 \textsubscript{(.65,.75)} &
            .82 \textsubscript{(.78,.85)} &
            .67 \textsubscript{(.62,.71)} &
            .74 \textsubscript{(.69,.79)} &
            .71 \textsubscript{(.66,.75)} &
            .82 \textsubscript{(.79,.86)} \\ \bottomrule
    \end{tabular}
    \end{adjustbox}
    \caption{Mean test set macro F1-score (and 95\% C.I.) across 5-fold cross validation. We compare BERT Base and Clinical BERT task models with a frozen (\raisebox{-0.4ex}{\footnotesize{\SnowflakeChevron}}) and unfrozen (\raisebox{-0.4ex}{\footnotesize{\FilledRainCloud}}) encoder. We also compare BERT Base and Clinical BERT task models with and without continued pretraining.}
    \label{tab:results}
\end{table*}

\section{Domain Adaptation Still Matters} \label{sec:exp}

Although regular expressions and hand-crafted rules can be used to extract clinical concepts with moderately high precision and better recall than diagnostic codes, they are poorly suited for inferring attributes for the extracted concepts. As an example, consider the task of inferring laterality or negation for the PDR span in Figure \ref{fig:annotationfile}. Contextual language models (CLMs) on the other hand have the ability to learn inter-token dependencies and, when explicit evidence is not available in the text, leverage prior knowledge to infer latent attributes \citep{devlin2018bert,liu2019linguistic}.  However, training a CLM from scratch typically requires a substantial amount of data, a constraint that is difficult to satisfy in many clinical NLP settings. 

To address data scarcity issues, we can instead use models pretrained on out-of-distribution data as a starting point and then fine-tune them for our target domain \citep{alsentzer2019publicly}. Nonetheless, maximizing performance under this regime is non-trivial. Which language model do we use as our foundation? Does the pretraining distribution matter? What about the vocabulary? We investigate these issues in the context of building our phenotyping system for diabetic eye disease.

\subsection{Do clinical LMs outperform non-clinical LMs in the presence of clinical data distribution shift?} \label{sec:exp-base-lm}

Prior studies have shown that LMs trained on clinical data may achieve better performance in downstream clinical tasks compared to LMs trained on non-clinical data \citep{lamproudis2022evaluating}. However, this effect is not consistent across tasks and datasets \citep{lewis2020pretrained,yang2022gatortron}, potentially due to differences between the pretraining and target distributions. In this first experiment, we ask whether an LM trained on data from a significantly different clinical setting than ophthalmology still transfers better to ophthalmology-related tasks than an LM trained with non-clinical data.

\textbf{Methods} Our primary task models consist of an encoder-style LM with a dense MLP output layer (see Figure \ref{fig:architecture} in Appendix \ref{apx:model-task}). As a naive baseline, we consider a majority classifier conditioned jointly on the target concept and token span (see Appendix \ref{apx:model-majority}). We conduct a 2 x 2 factorial experiment. As the first factor, we compare downstream task performance achieved using a general purpose LM  -- BERT Base (Cased) \citep{devlin2018bert} -- with the performance achieved using a clinical LM trained on notes from an ICU setting -- Clinical BERT \citep{alsentzer2019publicly}. As the second factor, we compare performance achieved with the encoder parameters frozen and unfrozen. Our evaluation metric is task-level macro F1-score. Additional experimental details are included in Appendix \ref{apx:expdesign}.

\textbf{Results} As shown in Table \ref{tab:results}, we do not observe any significant difference in performance between task models using BERT Base and Clinical BERT.\footnote{Our discussion and test statistics are based on average task performance. We use paired t-tests with a significance level of 0.05. Additional results are included in our digital supplement.\footnotemark[3]} Clinical BERT models achieve slightly higher average performance than BERT Base models when the encoder is frozen, but actually fall below BERT Base models when the encoder is unfrozen. This trend may suggest that advantages of pretraining with out-of-distribution clinical data are nullified once tuning models to a new clinical data distribution. On average, when the encoder is frozen, neither the Clinical BERT (t(69)=1.858, p=0.067) nor BERT Base (t(69)=1.530, p=0.131) task models significantly outperform the majority classifier baseline. In comparison, when the encoder is unfrozen, both Clinical BERT and BERT Base task models significantly outperform the majority classifier baseline and their frozen counterparts (p<0.001). Putting these results together, we note that task fine-tuning mitigates issues related to clinical data distribution shift. 


\subsection{Is task fine-tuning sufficient for adapting LMs to a new clinical data distribution?} \label{sec:exp-pretraining}

In low-resource settings, supervised task fine-tuning can be a sub-optimal method of LM transfer due to overfitting \citep{griesshaber2020fine,tinn2023fine}. Provided a sufficient corpus of text from a downstream task's domain, continued pretraining of the LM in a self-supervised manner can mitigate this risk \citep{gururangan2020don,deryshould}. In this experiment, we ask whether continued pretraining improves generalization beyond what is achieved via task fine-tuning.

\textbf{Methods} We use notes from all patients not in the annotated dataset to continue pretraining the BERT Base and Clinical BERT language models. Using each model's respective tokenizer, this amounts to approximately 192M tokens over 1.8M sequences (128 token max length).\footnote{BERT Base (Cased) and Clinical BERT use the same vocabulary, but their tokenizers have learned slightly different word-piece splitting criteria.} We train the language model for 16,500 steps using the standard masked language modeling objective \citep{devlin2018bert}, allowing for early stopping if validation loss starts increasing, and use the final checkpoint as the encoder in our task models. As before, we compare task performance with the encoder frozen and unfrozen. Additional training details are included in Appendix \ref{apx:model-lm}.

\textbf{Results} Continued pretraining leads to an additional, significant improvement in downstream task performance in each of the four settings considered in \S \ref{sec:exp-base-lm} (i.e., \{BERT Base, Clinical BERT\} $\times$ \{Frozen Encoder, Unfrozen Encoder\}). As shown in the bottom row of Table \ref{tab:results}, continued pretraining improves average macro F1 score for the BERT Base and Clinical BERT task models with a frozen encoder by 0.05 and 0.04, respectively. The effect of continued pretraining is more pronounced when we unfreeze the encoder, with the BERT Base and Clinical BERT task models achieving an average increase of 0.07 and 0.08 in macro F1 score, respectively, over their counterparts that did not undergo continued pretraining. Overall, our results indicate that both task fine-tuning \emph{and} continued pretraining are critical for maximizing downstream task performance. Furthermore, for our dataset, whether these procedures are applied to an encoder pretrained initially on clinical or non-clinical data does not make a difference in downstream performance.


\begin{figure}
    \centering
    \includegraphics[width=\linewidth]{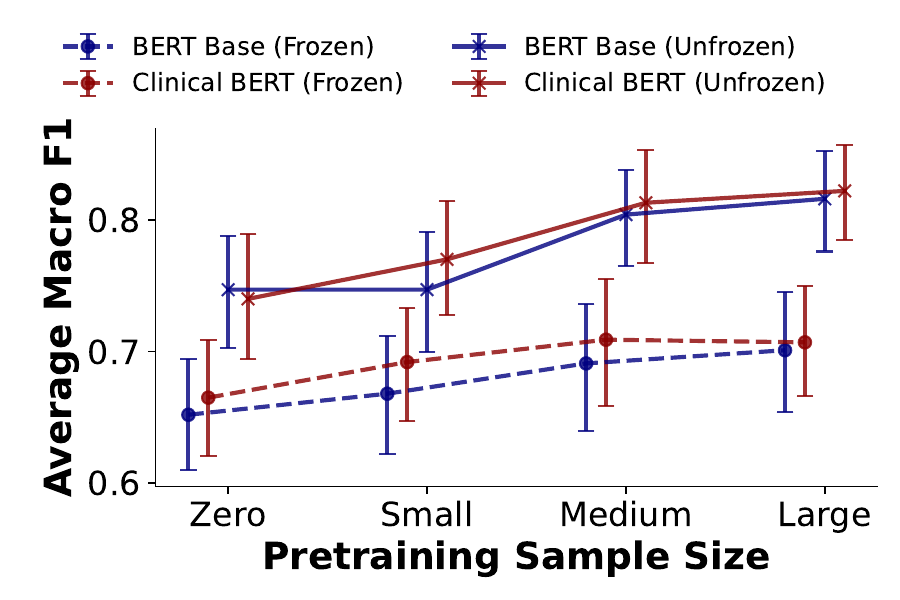}
    \caption{Average task performance as a function of pretraining sample size. Clinical BERT performs slightly better than BERT Base with little to no pretraining.}
    \label{fig:efficiency}
\end{figure}

\subsection{Are LMs pretrained on clinical data more efficient than LMs trained on non-clinical data in low-data regimes?} \label{sec:exp-size}

We find ourselves in a fortunate position, having access to a non-trivial amount of electronic medical records from our target clinical domain and reasonable compute infrastructure. If clinical language models don't significantly outperform non-clinical language models in the presence of these resources, then perhaps they require fewer resources to achieve equivalent levels of performance.

\textbf{Methods} We continue pretraining the BERT Base and Clinical BERT models using the same procedure outlined in \S \ref{sec:exp-pretraining}. However, we now consider 2 smaller subsets of the pretraining dataset -- $N$=1,024 (\texttt{Small}) and $N$=16,384 (\texttt{Medium}). As before, we continue pretraining for a maximum of 16,500 steps and use early stopping to prevent overfitting. We evaluate overall efficiency of the two LMs on the basis of sample efficiency -- downstream performance as a function of pretraining dataset size -- and compute efficiency -- the number of updates required before the stopping criteria is met. We contextualize performance against prior results -- no continued pretraining (\texttt{Zero}) and pretraining with all available data (\texttt{Large}).

\textbf{Results} Average task performance as a function of the pretraining dataset size is visualized in Figure \ref{fig:efficiency} and task-specific performance is included in Table \ref{tab:performance-size} of Appendix \ref{apx:performance-task-specific}. For 3 out of 4 settings, continued pretraining does not significantly increase downstream performance until utilizing the \texttt{Medium}-sized subset. The unfrozen, Clinical BERT task model is the sole exception, with continued pretraining on the \texttt{Small} subset causing a statistically significant improvement in performance (t(69)=2.582, p=0.012). That said, the frozen, Clinical BERT task model also seems to take advantage of the \texttt{Small} pretraining sample slightly better than the frozen, BERT Base model. The frozen, Clinical BERT task model nearly achieves a significant average improvement in macro F1 of 0.02 with the \texttt{Small} subset (t(69)=1.954, p=0.055), compared to the frozen, BERT Base task model's improvement in macro F1 of 0.01 with the \texttt{Small} subset (t(69)=1.136, p=0.260).


We note that both models approximately match the \texttt{Large} (full-dataset) pretraining performance using a fraction of the data (\texttt{Medium}). Both models trained on the \texttt{Small} subset reach stopping criteria at the same point -- 2300 steps -- while the Clinical BERT model reaches its stopping criteria 600 steps earlier than BERT Base on the \texttt{Medium} subset (6800 vs. 6200). Directionally, the Clinical BERT models appear to be more efficient in low-data regimes than BERT Base models. However, the differences are still relatively small, limiting the strength of conclusions we can draw.


\subsection{Can we ignore out-of-domain pretraining entirely?} \label{sec:exp-scratch}

We've seen that continued pretraining is critical to maximizing downstream performance. At the same time, the difference in performance when initializing with a clinical LM instead of a non-clinical LM has been marginal at best. It is natural to ask whether the initial pretraining phase was even necessary. Can we achieve the same downstream performance when training an LM on our dataset entirely from scratch? Doing so would remove an external dependency that could create non-trivial legal and robustness challenges upon deployment.

\textbf{Methods} We run the same pretraining procedure as above, but we now initialize the BERT encoder randomly instead of from an external checkpoint. We train one model using the BERT Base tokenizer, and one model using a tokenizer trained on our pretraining dataset. We compare downstream task performance achieved by the two models to each other and to the models initialized from an existing checkpoint. 

\begin{table}[t]
    \centering
    \begin{adjustbox}{width=\linewidth}
    \begin{tabular}{l c c c} \toprule
         \textbf{Initialization} & 
            \textbf{Tokenizer} &
            \raisebox{-0.5ex}{\SnowflakeChevron} & 
            \raisebox{-0.5ex}{\FilledRainCloud} \\ \midrule
         BERT Base & 
            BERT Base & 
            .70 \textsubscript{(.66,.75)} & 
            .82 \textsubscript{(.78,.85)} \\
         Random & 
            BERT Base & 
            .71 \textsubscript{(.66,.75)} & 
            .77 \textsubscript{(.73,.81)} \\
         Random & 
            Learned & 
            .71 \textsubscript{(.67,.76)} & 
            .81 \textsubscript{(.78,.84)}  \\ \bottomrule
    \end{tabular}
    \end{adjustbox}
    \caption{Average task performance for each model after pretraining on our dataset. We compare models without (\raisebox{-0.5ex}{\SnowflakeChevron}) and with task fine-tuning (\raisebox{-0.5ex}{\FilledRainCloud}).}
    \label{tab:results-scratch}
\end{table}

\textbf{Results} We report average task performance in Table \ref{tab:results-scratch} and include task-specific performance in Table \ref{tab:performance-initialization} of Appendix \ref{apx:performance-task-specific}. Task models using encoders initialized with random weights achieve roughly equivalent performance as those initialized from existing checkpoints. The domain-specific vocabulary magnifies the efficacy of task fine-tuning, enabling the randomly initialized LM to outperform the LMs pretrained first on other data distributions for some tasks. Once again, we observe that out-of-distribution clinical pretraining provides little-to-no benefit for our target domain.


\section{Discussion} \label{sec:discussion}

Within each of the experiments above, we find our clinical language models only match (or in some cases fall behind) their non-clinical counterparts. Moreover, we see that both task fine-tuning and continued pretraining using data from our specialized clinical domain are necessary to maximize downstream performance. These results align with and augment existing work that highlights shortcomings of out-of-domain clinical pretraining \citep{ranti2020utility,lin2020does,ji2021does,harrigian2023characterization}. At the same time, they temper the generality of the claim from \citet{lehman2023we} that language models pretrained on clinical data are necessary for clinical NLP tasks. Our results suggest it is more appropriate to say \emph{domain-specific} clinical language models are still necessary for clinical NLP tasks.

Accurately portraying the limits of clinical language models can have a meaningful impact when designing systems for novel clinical use cases such as our own. Cost, compute, and data privacy constraints already limit the breadth of parameters we can consider when developing a new clinical system. Choosing a sub-optimal pretrained language model as the system's foundation can introduce a performance ceiling before the first experiment is even executed. Notably, this decision is becoming increasingly difficult, with it now common for language models boasting improved performance along various task axes to be released daily \citep{FLI2023,alaga2023coordinated}.




Where clinical language models will ultimately reside amongst this deluge of resources remains uncertain. Most state of the art LMs draw performance not only from their complexity (i.e., \# parameters), but also their massive training datasets \citep{chung2022scaling,touvron2023llama}. Even if clinical data becomes easier to obtain, general text corpora from the internet will remain orders of magnitude larger than clinical corpora. Moreover, our experimental results suggest that in the presence of distributional shift, some form of domain adaptation (e.g., continued pretraining) is necessary to maximize an underlying clinical LM's utility, regardless of its pretraining data source. 

Together, these observations raise an important and timely question: should we focus on training general-purpose clinical LMs, or extracting performance from larger, non-clinical LMs via domain adaptation? This study has only evaluated two, albeit two widely-used, LMs, and is unable to answer this question. However, various other studies have already demonstrated that large, non-clinical LMs such as GPT-3 contain non-trivial amounts of medical knowledge and can perform well across several biomedical tasks \citep{agrawal2022large,singhal2023large,nori2023capabilities}. What remains to be seen is whether these levels of performance extend beyond the relatively small and homogeneous pool of clinical NLP benchmarks to novel domains.



\section*{Ethics Statement} \label{sec:ethics}

Our study involves the analysis of sensitive medical information from real patients. As such, our work is subject to appropriate Health Insurance Portability and Accountability Act (HIPAA) regulations and additional privacy policies set forth by our institution (e.g., compute environment restrictions, patient limits during annotation). We are currently unable to release models due to the risk of personal health information (PHI) leakage \citep{lukas2023analyzing}. All research was approved by our institutional review board (IRB) before its start and adhered to the tenets of the Declarations of Helsinki.



\section*{Limitations} \label{sec:limitations}


\textbf{Data} There are four data-related limitations in our study. First, our dataset is drawn from a single, academic hospital system in a mid-sized U.S. city. Our patient population differs from other geographic locations, as does the hospital system's policies, documentation practices, and clinical priorities. Second, although annotations in our dataset were agreed upon by two domain experts, it is still possible that application of our ontology is imperfect. Third, we note that attributes which could not be confidently labeled by annotators were treated as missing data; it is possible that our dataset is biased such that it contains ``easier'' examples. And fourth, our dataset is relatively small by contemporary standards and exhibits significant class/concept imbalance. Confidence intervals in Table \ref{tab:results} reflect uncertainty that arises due to this limitation.

\textbf{Experimental Design} There are three major limitations with our experimental design to remain cognizant of when interpreting results. First, the concept and class imbalance issues mentioned above necessitated that we consolidate certain ontology attributes and classes. Although this was accomplished with guidance from domain experts, it is possible that alternative groupings would have maintained more (or less) statistical signal. Second, as is common in empirical studies, computational constraints limited the breadth of hyperparameters which were explored to optimize performance. This is especially pertinent for the pretraining results, in which only a single LM could be trained due to computational expense. Finally, we note that our study only focused on two base LMs which have the same transformer architecture and vocabulary. Alternative language models with different architectures \citep{lehman2023we} and pretraining data \citep{lee2020biobert,yang2022gatortron} may have yielded different outcomes.

With respect to the latter, while it is true that BERT language models are comparatively small in the modern language modeling landscape, we argue they still are capable of providing insight regarding the value of out-of-domain clinical pretraining. BERT-style models have outperformed alternative architectures in several clinical NLP tasks and across datasets, even those which are significantly larger \citep{agrawal2022large,gutierrez2022thinking,lu2022clinicalt5,lehman2023we,labrak2023zero,rehana2023evaluation}. Furthermore, clinical datasets such as ours are generally orders of magnitude smaller than non-clinical datasets and may not even support the training of larger architectures \citep{spasic2020clinical,touvron2023llama,wornow2023shaky}. Our intention is not to make sweeping claims regarding clinical LMs, but rather to encourage further exploration of clinical LM shortcomings and the heterogeneous nature of clinical language domains.


\acks{Dr. Cai was funded by a Career Development Award from the Research to Prevent Blindness organization, and by a K23 award from the National Institutes of Health (NIH) and National Eye Institute (NEI) (Award No. K23EY033440). Dr. Cai is the Jonathan and Marcia Javitt Rising Professor of Ophthalmology.}


\bibliography{jmlr-sample}


\appendix

\section{Abbreviations} \label{apx:abbreviations}

Enumerated below is a list of useful clinical abbreviations used throughout our study.

\begin{itemize}[label={},leftmargin=0em]
    \setlength\itemsep{0em}
    \item \textbf{DR}: Diabetic Retinopathy
    \item \textbf{PDR}: Proliferative Diabetic Retinopathy 
    \item \textbf{NPDR}: Non-proliferative Diabetic Retinopathy 
    \item \textbf{HR-PDR}: High-Risk PDR
    \item \textbf{NHR-PDR}: Non-High-Risk PDR
    \item \textbf{NV}: Neovascularization
    \item \textbf{NVD}: Neovascularization of the Disc
    \item \textbf{NVE}: Neovascularization of the Retina Elsewhere 
    \item \textbf{AMD}: Age-related Macular Degnereration
    \item \textbf{NVI}: Neovascularization of the Iris
    \item \textbf{ME}: Macular Edema
    \item \textbf{DME}: Diabetic Macular Edema
    \item \textbf{CI-DME}: Center Involved DME
    \item \textbf{CS-DME}: Clinically Significant DME
    \item \textbf{CME}: Cystoid Macular Edema
    \item \textbf{VH}: Vitreous Hemorrhage
    \item \textbf{NVG}: Neovascular Glaucoma
    \item \textbf{RD}: Retinal Detachment
    \item \textbf{TRD}: Traction Retinal Detachment
    \item \textbf{RRD}: Rhegmatogenous Retinal Detachment
    \item \textbf{Anti-VEGF}: Anti–vascular Endothelial Growth Factor Therapy
    \item \textbf{PRP}: Panretinal Photocoagulation
    \item \textbf{DM}: Diabetes Mellitus
    \item \textbf{OS}: Oculus Sinister (Left Eye)
    \item \textbf{OD}: Oculus Dexter (Right Eye)
    \item \textbf{OU}: Oculus Uterque (Both Eyes)
\end{itemize}

\section{Data} \label{apx:data}

\subsection{Concept Ontology} \label{apx:data-ontology}

We present our clinical concept ontology in Table \ref{tab:ontology}. Groups (e.g., Retina Conditions, Comorbidities) are included for clarity, but not used explicitly in our study. The ontology was fine-tuned over multiple iterations of pilot annotation experiments to balance label utility with the cognitive load required by annotators to apply the ontology consistently.

\subsection{Annotation} \label{apx:data-process}

\begin{table}[t]
    \centering
    \small
    \setlength{\tabcolsep}{5pt}
    \begin{tabular}{lccccc} \toprule
    \multicolumn{1}{c}{\textbf{Concept}} & 
            \rotatebox[origin=l]{90}{\textbf{Valid}} & 
            \rotatebox[origin=l]{90}{\textbf{Invalid}} & 
            \rotatebox[origin=l]{90}{\textbf{Laterality}} &
            \rotatebox[origin=l]{90}{\begin{tabular}{@{}c@{}}\textbf{Severity/}\\\textbf{Type}\end{tabular}} &
            \rotatebox[origin=l]{90}{\textbf{Temporality}} \\ \midrule
    DR (Generic)                    &    455 &        3 &         446 &              0 &          425 \\
    NPDR                            &    194 &        0 &         189 &            165 &          187 \\
    PDR                             &    209 &        0 &         203 &            170 &          189 \\
    NV                              &    360 &        1 &         354 &            332 &          355 \\
    ME                              &    896 &       12 &         848 &            845 &          777 \\
    VH                              &    266 &        1 &         264 &              0 &          253 \\
    RD                              &    275 &        1 &         258 &            145 &          263 \\
    NVG                             &    369 &        0 &         356 &              0 &          355 \\
    Anti-VEGF                       &    508 &        0 &         501 &              0 &          465 \\
    PRP                             &    211 &        0 &         205 &              0 &          207 \\
    Focal Laser                     &     16 &        0 &          16 &              0 &           16 \\
    Other Injections                &     16 &        0 &          16 &              0 &           16 \\
    Retina Surgery                  &    226 &       37 &         212 &            118 &          198 \\
    NVG Surgery                     &     70 &        4 &          63 &             63 &           65 \\
    Diabetes                        &   1,045 &        0 &           0 &            785 &         1,029 \\
    Nephropathy                     &    340 &        0 &           0 &              0 &          337 \\
    Neuropathy                      &    351 &        0 &           0 &              0 &          345 \\
    Heart Attack                    &    342 &        1 &           0 &              0 &          336 \\
    Stroke                          &    355 &        1 &           0 &              0 &          351 \\
    \midrule
    \textbf{Total}                       & 6,504 & 61 & 3,931 & 2,623 & 6,169 \\
    \bottomrule
    \end{tabular}
    \caption{Distribution of spans and attribute labels for the 19 clinical concepts in our ontology.}
    \label{tab:labeldist}
\end{table}

Two annotators with clinical expertise in diabetic eye disease (a PGY-4 ophthalmology resident and a licensed optometrist) were responsible for all annotations. Efforts were overseen by a board-certified ophthalmologist, an author of this study, who designed the concept ontology. Both annotators had access to an annotation guide containing edge-case examples and other rules for applying the concept ontology to the free text notes. 

Annotation was completed in a secure remote desktop environment using a Microsoft Excel workbook outfitted with conditional data validation cells and custom text formatting (see Figure \ref{fig:annotationfile}). Notes were shared with us in sanitized CSV files, thus necessitating the use of additional rules to format the text to be more readable (e.g., line-breaks, removal of tables containing lab results). Metadata for the encounter (i.e., ICD-10 codes) was included in the formatted note and annotated in the same manner as free text.

Annotation was completed during two phases, each consisting of multiple rounds. During the first phase, a random sample of 236 notes from 139 patients containing at least one concept span were labeled. During the first round of the first phase, annotators labeled 3,013 concept-spans independently. During the second round, annotators were asked to independently relabel all concept spans which contained disagreement during the first round (869 spans). During the third and final round, annotators resolved any remaining disagreements via discussion (501 spans).

A review of the label distribution after the first phase of annotation suggested that more annotation would be necessary to address severe class imbalance. For the second phase of annotation, a random sample of 500 notes from 209 patients (different from those in the first phase) where labeled. During the first round of the second phase, annotators labeled 3,552 concept-spans independently. Rather than including another round of independent review as was done during the first phase, annotators met immediately after the first round to resolve any disagreement via discussion (790 spans).

\subsection{Concept Extraction} \label{apx:data-extract}

We ran multiple pilot experiments to optimize the annotation procedure described above. At the beginning of the study, our institution hosted internally-developed, HIPAA-compliant software which supported span-level extraction and annotation.\footnote{\url{https://github.com/JHUAPL/PINE}} Unfortunately, this service was discontinued after a single pilot experiment, in turn requiring us to seek alternative annotation protocols. External software could not be deployed easily while maintaining patient privacy. At the same time, our pilot experiment suggested that annotators would struggle to identify relevant concept spans over the large label space in an efficient manner without compromising accuracy. Together, these circumstances motivated us to develop high-recall regular expressions which could be used to automatically identify relevant concept spans which would then be manually reviewed and further annotated by annotators.

We used labels generated from the HIPAA-compliant annotation software during the aforementioned pilot experiment in combination with domain knowledge from members of our team to construct the base set of regular expressions. We then augmented this set with ICD-10 codes based on data from an online resource.\footnote{\url{https://www.icd10data.com}} ICD-10 code matches were only included for annotation if their associated clinical concepts were not already found in the note free text; this decision was made to limit redundant annotation efforts. In general, regular expressions were designed to capture the most generic manifestation of each concept (e.g., ``NPDR'' instead of ``Severe NPDR''). We provide our expressions and parsing logic in the supplemental material.\footnote{\url{https://github.com/kharrigian/ml4h-clinical-bert}}

\begin{table}[t]
    \centering
    \small
    \begin{tabular}{lrrr} \toprule
    \multicolumn{1}{c}{\begin{tabular}{@{}c@{}}\textbf{Clinical}\\\textbf{Concept}\end{tabular}} & \multicolumn{1}{c}{\textbf{ICD-10}} & \multicolumn{1}{c}{$\boldsymbol{\cap}$} & \multicolumn{1}{c}{\begin{tabular}{@{}c@{}}\textbf{Text}\end{tabular}} \\ \midrule
    A1 - DR (General)                    &    37,743 &       85,878 &    55,848 \\
    A2 - NPDR                            &    24,970 &       28,203 &    14,594 \\
    A3 - PDR                             &    14,217 &       32,212 &    13,038 \\
    A4 - NV                              &     7,624 &       14,945 &    58,970 \\
    B1 - ME                              &    37,081 &       83,852 &    50,719 \\
    C1 - VH                              &     3,569 &        4,882 &    28,691 \\
    C2 - RD                              &     4,337 &       15,279 &    70,781 \\
    C3 - NVG                             &   155,429 &        6,127 &     3,418 \\
    D1 - Anti-VEGF                       &         0 &            0 &    93,038 \\
    D2 - PRP                             &         0 &            0 &    41,531 \\
    D3 - Focal Grid Laser                &         0 &            0 &     7,339 \\
    D4 - Other Injections                &         0 &            0 &     8,230 \\
    E1 - Retina Surgery                  &         0 &            0 &    90,680 \\
    E2 - NVG Surgery                     &        19 &           11 &    34,980 \\
    F1 - Diabetes Mellitus               &    44,019 &      165,813 &    54,499 \\
    G1 - Nephropathy                     &     3,073 &           33 &     1,941 \\
    G2 - Neuropathy                      &     5,615 &          363 &     5,997 \\
    G3 - Heart Attack                    &        56 &            3 &     4,796 \\
    G4 - Stroke                          &       783 &          921 &    10,452 \\
    \bottomrule
    \end{tabular}
    \caption{Number of notes containing at least one regular expression match for each clinical concept in our ontology, faceted by location of the match. Free-text search improves recall of relevant clinical concepts over using ICD-10 codes alone.}
    \label{tab:texticdcompare}
\end{table}

The distribution of concept-spans and attribute labels is provided in Table \ref{tab:labeldist}. False positives are those which were marked as “Invalid” spans by annotators. An example of an invalid span is provided in Figure \ref{fig:annotationfile} -- ``eylea left eye after vacation to <<MI>>'' -- where the proposed span ``<<MI>>'' refers to the geographic location of Michigan, not Myocardial Infarction. As seen in the table, precision of the regular expressions varies as a function of clinical concept. The lowest level of precision (0.86) occurs for Retina Surgery, with mentions of `laser' being the largest source of error. The overall precision across all concepts spans was greater than 0.99. 

We initially planned to train classifiers to infer span validity for all clinical concepts. However, as shown in Table \ref{tab:labeldist}, only two concepts – Macular Edema, Retina Surgery – had enough invalid spans to support/warrant this modeling. Spans which were marked as invalid by annotators were not used for training the attribute classification models; they were only used for training the Macular Edema and Retina Surgery span validity classifiers.

The aforementioned limitations regarding span-level annotation made it difficult to empirically estimate recall. However, we note that this system's goal is to improve recall of clinical concepts over what is possible through diagnostic codes alone. Accordingly, in Table \ref{tab:texticdcompare}, we show the distribution of concept matches in the entire 692,486 note dataset broken down by location of the match. The right-most column is most important, indicating the number of additional notes which were identified as relevant to our clinical use case by examining the free-text. Some relevant concepts (e.g., surgical procedures) cannot be found by examining ICD-10 codes, while others receive significantly higher coverage by looking at the free text. Besides improving recall as desired, the free-text search allows us to independently characterize multiple instances of the same concept (e.g., Mild NPDR $\rightarrow$ Moderate NPDR).

\subsection{Task Consolidation} \label{apx:data-tasks}

The annotated dataset exhibits sparsity for some of the clinical concepts and heavy class imbalance for the majority of attributes. Rather than train independent models for each concept-attribute pair in our dataset, we draw statistical strength from the overlapping label space by modeling multiple concepts jointly and merging attribute labels with high semantic similarity in the diabetic eye disease use case. For example, we model temporality for the retina conditions and their complications together, while also remapping the original temporality attribute labels into a binary output set indicating whether the concept is present (or relevant) at the time of the encounter.

The consolidated summary of classification tasks is presented in Table \ref{tab:mapping}. In addition to the remapped labels, the reader may also note the exclusion of certain concept-attribute pairs altogether. We decide not to model temporality for \texttt{F1 - Diabetes Mellitus} due to extreme label imbalance -- all but 9 spans are Active, and these 9 spans are concentrated in only 5 patients. We also decide to model span validity (i.e., whether the regular expression match was correct) for only \texttt{B1 - Macular Edema} and \texttt{E1 - Retina Surgery} because the number of false positives for the remaining concepts was small and unlikely to yield a useful model. The label distribution for each of the 14 tasks in provided in Table \ref{tab:labeldistfull}.

\begin{table*}[t]
    \centering
    \small
    \begin{adjustbox}{width=\linewidth}
    \begin{tabular}{c c c c c} \toprule
        \textbf{Group} & \textbf{Concept} & \textbf{Laterality} & \textbf{Temporality} & \textbf{Severity/Type} \\ \midrule
        \multirow{12}{*}{\begin{tabular}{@{}c@{}}Retina\\Conditions\end{tabular}} & 
            A1 - DR (General) & 
            OS, OD, OU & 
            Active, History of & 
            -- \\ \cmidrule(l){2-5}
        & 
            A2 - NPDR & 
            OS, OD, OU & 
            Active, History of & 
            \begin{tabular}{@{}c@{}}Mild, Mild-Moderate, Moderate,\\Moderate-Severe, Severe\end{tabular} \\ \cmidrule(l){2-5}
        & 
            A3 - PDR & 
            OS, OD, OU & 
            Active, History of & 
            NHR-PDR, HR-PDR  \\ \cmidrule(l){2-5}
        & 
            A4 - NV & 
            OS, OD, OU & 
            Active, Resolved &
            \begin{tabular}{@{}c@{}}Iris, Iris + NVD and/or NVE,\\NVD, NVE, NVD/NVE,\\AMD, Other\end{tabular}  \\ \cmidrule(l){2-5}
         & 
            B1 - ME & 
            OS, OD, OU & 
            Active, History of & 
            \begin{tabular}{@{}c@{}}DME, CI-DME, Non-CI-DME,\\CS-DME, Non-CS-DME, CME,\\AMD, Other\end{tabular}  \\ \midrule
        \multirow{5}{*}{\begin{tabular}{@{}c@{}}Complications\\of Retina\\Conditions\end{tabular}} & 
            C1 - VH & 
            OS, OD, OU & 
            \begin{tabular}{@{}c@{}}Active, History of,\\Resolving, Resolved\end{tabular} &
            --  \\ \cmidrule(l){2-5}
        & 
            C2 - RD & 
            OS, OD, OU & 
            Active, History of & 
            \begin{tabular}{@{}c@{}}RRD, TRD, Serous,\\Combined RRD/TRD\end{tabular}  \\ \cmidrule(l){2-5}
        & 
            C3 - NVG & 
            OS, OD, OU & 
            Present, Not Present & 
            --  \\ \midrule
        \multirow{6}{*}{\begin{tabular}{@{}c@{}}Treatment\\(Procedure)\end{tabular}} & 
            D1 - Anti-VEGF & 
            OS, OD, OU &
            \begin{tabular}{@{}c@{}}History of, Performed Today,\\Recommended, Considering\end{tabular} & 
            --  \\ \cmidrule(l){2-5}
        & 
            D2 - PRP & 
            OS, OD, OU & 
            \begin{tabular}{@{}c@{}}History of, Performed Today,\\Recommended, Considering\end{tabular} & 
            -- \\ \cmidrule(l){2-5}
        & 
            \begin{tabular}{@{}c@{}}D3 - Focal\\Grid Laser\end{tabular} & 
            OS, OD, OU & 
            \begin{tabular}{@{}c@{}}History of, Performed Today,\\Recommended, Considering\end{tabular} & 
            -- \\ \cmidrule(l){2-5}
        & \begin{tabular}{@{}c@{}}D4 - Other\\Intravitreal\\Injections \end{tabular} & 
            OS, OD, OU & 
            \begin{tabular}{@{}c@{}}History of, Performed Today,\\Recommended, Considering\end{tabular} & 
            --  \\ \midrule
        \multirow{4}{*}{\begin{tabular}{@{}c@{}}Treatment\\(Surgery)\end{tabular}} & 
            E1 - Retina  Surgery & 
            OS, OD, OU & 
            \begin{tabular}{@{}c@{}}History of, Performed Today,\\Recommended, Considering\end{tabular} &  
            \begin{tabular}{@{}c@{}}Indication VH,\\Indication RD\end{tabular} \\ \cmidrule(l){2-5}
        & 
            E2 - NVG Surgery & 
            OS, OD, OU & 
            \begin{tabular}{@{}c@{}}History of, Performed Today,\\Recommended, Considering\end{tabular} & 
            \begin{tabular}{@{}c@{}}Tube, Trab,\\MIGS\end{tabular}  \\ \midrule
        Comorbidities & 
            \begin{tabular}{@{}c@{}}F1 - Diabetes\\Mellitus\end{tabular} & 
            -- & 
            Active, Resolved & 
            \begin{tabular}{@{}c@{}}Type I, Type II,\\Gestational, Other\end{tabular}\\ \midrule
        \multirow{6}{*}{\begin{tabular}{@{}c@{}}Complications\\of Diabetes\\Mellitus\end{tabular}} & 
            G1 - Nephropathy & 
            -- & 
            Present, Not Present & 
            -- \\ \cmidrule(l){2-5}
        & 
            G2 - Neuropathy & 
            --  & 
            Present, Not Present & 
            -- \\ \cmidrule(l){2-5}
        & 
            G3 - Heart Attack & 
            -- & 
            History of, No History of & 
            -- \\ \cmidrule(l){2-5}
        & 
            G4 - Stroke &
            -- & 
            History of, No History of & 
            -- \\ \bottomrule
    \end{tabular}
    \end{adjustbox}
    \caption{Ontology of concepts related to diabetic eye disease. We include definitions for all abbreviations in Appendix \ref{apx:abbreviations}. Where appropriate, temporality classes can be negated (e.g.,  Negation + History of = No History of).}
    \label{tab:ontology}
\end{table*}

\begin{table*}[t]
    \centering
    \small
    \begin{adjustbox}{width=0.8\linewidth}
    \begin{tabular}{c c c c c} \toprule
     \textbf{Attribute} & 
        \textbf{Concepts} & 
        \textbf{Concept IDs} & 
            \multicolumn{1}{c}{\textbf{Consolidated}} & 
            \multicolumn{1}{c}{\textbf{Original}} \\ \midrule
     \multirow{13}{*}{Temporality} & 
        \multirow{4}{*}{Retina Conditions} & 
        \multirow{4}{*}{\begin{tabular}{@{}c@{}}A1, A2, A3, A4,\\B1, C1, C2, C3\end{tabular}} & 
            Present & 
            \begin{tabular}{@{}c@{}}Active, $\neg$ Resolved,\\Resolving, Present\end{tabular} \\ \cmidrule(l){4-5}
     & 
        & 
        & 
            Not Present & 
            \begin{tabular}{@{}c@{}}History of, $\neg$ Active,\\$\neg$ History of, Resolved\end{tabular} \\  \cmidrule(l){2-5}
    &
        \multirow{4}{*}{\begin{tabular}{@{}c@{}}Complications of\\Diabetes Mellitus\end{tabular}} &
        \multirow{4}{*}{\begin{tabular}{@{}c@{}}G1, G2, G3, G4\end{tabular}} &
            Present &
            \begin{tabular}{@{}c@{}}Present, History of,\\Active, $\neg$ Resolved\end{tabular} \\ \cmidrule(l){4-5}
    &   
        &
        &
            Not Present &
            \begin{tabular}{@{}c@{}}Not Present, No History of,\\$\neg$ Active, Resolved\end{tabular} \\ \cmidrule(l){2-5}
    &
        \multirow{5}{*}{Treatments} &
        \multirow{5}{*}{\begin{tabular}{@{}c@{}}D1, D2, D3,\\D4, E1, E2\end{tabular}} &
            History of &
            \multicolumn{1}{c}{--} \\ \cmidrule{4-5}
    &
        &
        &
            Performed Today &
            \multicolumn{1}{c}{--} \\ \cmidrule{4-5}
    &
        &
        &
            Discussed &
            \begin{tabular}{@{}c@{}}Recommended, Considering,\\$\neg$ History of, $\neg$ Performed Today,\\$\neg$ Recommended, $\neg$ Considering\end{tabular} \\ \midrule
    \multirow{4}{*}{Laterality} &
        \multirow{4}{*}{All} &
        \multirow{4}{*}{\begin{tabular}{@{}c@{}}A1, A2, A3, A4,\\B1, C1, C2, C3,\\D1, D2, D3, D4,\\E1, E2\end{tabular}} &
            OS & 
            \multicolumn{1}{c}{--} \\ \cmidrule(l){4-5}
    &
        &
        &
            OD &
            \multicolumn{1}{c}{--} \\ \cmidrule(l){4-5}
    &
        &
        &
            OU &
            \multicolumn{1}{c}{--} \\ \midrule
    \multirow{27}{*}{Type} &
        \multirow{7}{*}{Neovascularization} & 
        \multirow{7}{*}{A4} &
            NVD and/or NVE &
            NVD, NVE, NVD/NVE \\ \cmidrule(l){4-5}
    &
        &
        &
            Iris + NVD and/or NVE &
            \multicolumn{1}{c}{--} \\ \cmidrule(l){4-5}
    &
        &
        &
            Iris &
            \multicolumn{1}{c}{--} \\ \cmidrule(l){4-5}
    &
        &
        &
            AMD &
            \multicolumn{1}{c}{--} \\ \cmidrule(l){4-5}
    &
        &
        &
            Other &
            \multicolumn{1}{c}{--} \\ \cmidrule(l){2-5}
     &
        \multirow{3}{*}{Macular Edema} & 
        \multirow{3}{*}{B1} &
            DME &
            \begin{tabular}{@{}c@{}}DME, CI-DME, CS-DME,\\Non-CI-DME, Non-CS-DME\end{tabular} \\ \cmidrule(l){4-5}
    &
        &
        &
            Other &
            CME, AMD, Other \\ \cmidrule(l){2-5}
    &
        \multirow{6}{*}{Retinal Detachment} & 
        \multirow{6}{*}{C2} &
            RRD &
            \multicolumn{1}{c}{--} \\ \cmidrule(l){4-5}
    &
        &
        &
            TRD &
            \multicolumn{1}{c}{--} \\ \cmidrule(l){4-5}
    &
        &
        &
            Combined RRD/TRD &
            \multicolumn{1}{c}{--} \\ \cmidrule(l){4-5}
    &
        &
        &
            Serous &
            \multicolumn{1}{c}{--} \\ \cmidrule(l){2-5}
    &
        \multirow{4}{*}{Diabetes Mellitus} & 
        \multirow{4}{*}{F1} &
            Type I &
            \multicolumn{1}{c}{--} \\ \cmidrule(l){4-5}
    &
        &
        &
            Type II &
            \multicolumn{1}{c}{--} \\ \cmidrule(l){4-5}
    &
        &
        &
            Other &
            Gestational, Other \\ \cmidrule(l){2-5}
    &
        \multirow{3}{*}{Retina Surgery} & 
        \multirow{3}{*}{E1} &
            Indication VH &
            \multicolumn{1}{c}{--} \\ \cmidrule(l){4-5}
    &
        &
        &
            Indication RD &
            \multicolumn{1}{c}{--} \\ \cmidrule(l){2-5}
    &
        \multirow{4}{*}{NVG Surgery} & 
        \multirow{4}{*}{E2} &
            Tube &
            \multicolumn{1}{c}{--} \\ \cmidrule(l){4-5}
    &
        &
        &
            Trab &
            \multicolumn{1}{c}{--} \\ \cmidrule(l){4-5}
    &
        &
        &
            MIGS &
            \multicolumn{1}{c}{--} \\ \midrule
    \multirow{7}{*}{Severity} &
        \multirow{4}{*}{NPDR} &
        \multirow{4}{*}{A2} &
            Mild &
            Mild \\ \cmidrule(l){4-5}
    &
        &
        &
            Moderate &
            Mild-Moderate, Moderate \\ \cmidrule(l){4-5}
    &
        &
        &
            Severe &
            Moderate-Severe, Severe \\ \cmidrule(l){2-5}
    &
        \multirow{2.5}{*}{PDR} &
        \multirow{2.5}{*}{A3} &
            HR-PDR &
            \multicolumn{1}{c}{--} \\ \cmidrule(l){4-5}
    &
        &
        &
            NHR-PDR &
            \multicolumn{1}{c}{--} \\ \midrule
    \multirow{5}{*}{\begin{tabular}{@{}c@{}}Span\\Validity\end{tabular}} &
        \multirow{2}{*}{Macular Edema} &
        \multirow{2}{*}{B1} &
            Valid &
            \multicolumn{1}{c}{--} \\ \cmidrule(l){4-5}
    &
        &
        &
            Invalid &
            \multicolumn{1}{c}{--} \\ \cmidrule(l){2-5}
     &
        \multirow{2}{*}{Retina Surgery} &
        \multirow{2}{*}{E1} &
            Valid &
            \multicolumn{1}{c}{--} \\ \cmidrule(l){4-5}
    &
        &
        &
            Invalid &
            \multicolumn{1}{c}{--} \\ \bottomrule
    \end{tabular}
    \end{adjustbox}
    \caption{Consolidation of our concept ontology into 14 classification tasks. The ($\neg$) symbol denotes negation.}
    \label{tab:mapping}
\end{table*}

\begin{table*}
\centering
\setlength{\tabcolsep}{1.5pt}
\small
\begin{adjustbox}{width=0.9\linewidth}
\renewcommand*\arraystretch{1.1}
\begin{tabular}{cccccccccccccccccccccc}
\toprule
              \multicolumn{3}{c}{\textbf{Concept ID} $\boldsymbol{\longrightarrow}$} & 
              \rotatebox[origin=c]{70}{\textbf{A1}} &
              \rotatebox[origin=c]{70}{\textbf{A2}} & 
              \rotatebox[origin=c]{70}{\textbf{A3}} & 
              \rotatebox[origin=c]{70}{\textbf{A4 }} & 
              \rotatebox[origin=c]{70}{\textbf{B1}} & 
              \rotatebox[origin=c]{70}{\textbf{C1}} & 
              \rotatebox[origin=c]{70}{\textbf{C2}} & 
              \rotatebox[origin=c]{70}{\textbf{C3}} & 
              \rotatebox[origin=c]{70}{\textbf{D1}} & 
              \rotatebox[origin=c]{70}{\textbf{D2}} & 
              \rotatebox[origin=c]{70}{\textbf{D3}} & 
              \rotatebox[origin=c]{70}{\textbf{D4}} & 
              \rotatebox[origin=c]{70}{\textbf{E1}} & 
              \rotatebox[origin=c]{70}{\textbf{E2}} & 
              \rotatebox[origin=c]{70}{\textbf{F1}} & 
              \rotatebox[origin=c]{70}{\textbf{G1}} & 
              \rotatebox[origin=c]{70}{\textbf{G2}} & 
              \rotatebox[origin=c]{70}{\textbf{G3}} & 
              \rotatebox[origin=c]{70}{\textbf{G4}} \\
\midrule
\multirow{10}{*}{\rotatebox[origin=l]{90}{\textbf{Temporality}}} & 
    \multirow{2.5}{*}{Retina} & 
    Not Present & 309 & 7 & 65 & 270 & 385 & 133 & 235 & 326 &  &  &  &  &  &  & &  &  &  &  \\ \cmidrule{4-11} 
&  & Present & 116 & 180 & 124 & 85 & 392 & 120 & 28 & 29 &  &  &  &   &  &   &   &    &   &     &   \\
\cmidrule{2-22}
& \multirow{2.5}{*}{\begin{tabular}{@{}c@{}}DM\\Complications\end{tabular}} & Not Present &     &           &          &         &         &         &         &          &  &          &         &          &       &    &          & 334 &  331 & 334 & 331 \\ \cmidrule{19-22}
&  & Present &     &           &          &         &         &         &         &          &  &          &         &          &       &    &          &  3 & 14 & 2 & 20 \\
\cmidrule{2-22}
& \multirow{4}{*}{Treatment} & History of &     &           &          &         &         &         &         &          &            265 &      180 &      14 &        7 &   139 & 50 &          &    &   &     &   \\ \cmidrule{12-17}
&  & No Action &     &           &          &         &         &         &         &          &            145 &       22 &       2 &        8 &    58 & 15 &          &    &   &     &             \\ \cmidrule{12-17}
&  & Performed &     &           &          &         &         &         &         &          &             55 &        5 &      0  &        1 &     1 & 0  &          &    &   &     &             \\
\midrule 
\multirow{4}{*}{\rotatebox[origin=l]{90}{\textbf{Laterality}}} & \multirow{4}{*}{All} & OD &  20 &        25 &       49 &      99 &     236 &      97 &      87 &       27 &            217 &       79 &       1 &        3 &   119 & 24 &          &    &   &     &             \\ \cmidrule{4-17}
&  & OS &  11 &        24 &       32 &     163 &     251 &     152 &      61 &       12 &            212 &       95 &      11 &       10 &    61 & 15 &          &    &   &     &   \\ \cmidrule{4-17}
&  & OU & 415 &       140 &      122 &      92 &     361 &      15 &     110 &      317 &             72 &       31 &       4 &        3 &    32 & 24 &          &    &   &     &  \\ \midrule 
\multirow{23}{*}{\rotatebox[origin=l]{90}{\textbf{Type}}} & \multirow{2.5}{*}{ME} & DME &     &           &          &         &     672 &         &         &          &  &          &         &          &       &    &          &    &   &     &             \\ \cmidrule{8-8}
&  & Other &     &           &          &         &     173 &         &         &          &  &          &         &          &       &    &          &    &   &     &             \\
\cmidrule{2-22}
& \multirow{2.5}{*}{RD} & RRD &     &           &          &         &         &         &      97 &          &  &          &         &          &       &    &          &    &   &     &             \\ \cmidrule{10-10}
&  & TRD &     &           &          &         &         &         &      48 &          &  &          &         &          &       &    &          &    &   &     &             \\
\cmidrule{2-22}
& \multirow{7}{*}{NV} & AMD &     &           &          &      27 &         &         &         &          &  &          &         &          &       &    &          &    &   &     &             \\ \cmidrule{7-7}
&  & Iris &     &           &          &      31 &         &         &         &          &  &          &         &          &       &    &          &    &   &     &             \\ \cmidrule{7-7}
&  & Iris + NVD/NVE &     &           &          &      39 &         &         &         &          &  &          &         &          &       &    &          &    &   &     &             \\ \cmidrule{7-7}
&  & NVD/NVE &     &           &          &     234 &         &         &         &          &  &          &         &          &       &    &          &    &   &     &             \\ \cmidrule{7-7}
&  & Other &     &           &          &       1 &         &         &         &          &  &          &         &          &       &    &          &    &   &     &             \\
\cmidrule{2-22}
& \multirow{4}{*}{DM} & Other &     &           &          &         &         &         &         &          &  &          &         &          &       &    &        9 &    &   &     &     \\ \cmidrule{18-18}
&  & Type 1 &     &           &          &         &         &         &         &          &  &          &         &          &       &    &      173 &    &   &     &             \\ \cmidrule{18-18}
&  & Type 2 &     &           &          &         &         &         &         &          &  &          &         &          &       &    &      603 &    &   &     &             \\
\cmidrule{2-22}
& \multirow{4}{*}{\begin{tabular}{@{}c@{}}NVG\\Surgery\end{tabular}} & MIGS &     &           &          &         &         &         &         &          &  &          &         &          &       &  2 &          &    &   &     &             \\ \cmidrule{17-17}
&  & Trab &     &           &          &         &         &         &         &          &  &          &         &          &       & 36 &          &    &   &     &             \\ \cmidrule{17-17}
&  & Tube &     &           &          &         &         &         &         &          &  &          &         &          &       & 25 &          &    &   &     &             \\
\cmidrule{2-22}
& \multirow{2.5}{*}{\begin{tabular}{@{}c@{}}Retina\\Surgery\end{tabular}} & Indication RD &     &           &          &         &         &         &         &          &  &          &         &          &    67 &    &          &    &   &     &             \\ \cmidrule{16-16}
&  & Indication VH &     &           &          &         &         &         &         &          &  &          &         &          &    51 &    &          &    &   &     &             \\
\midrule 
\multirow{6.5}{*}{\rotatebox[origin=l]{90}{\textbf{Severity}}} & \multirow{4}{*}{NPDR} & Mild &     &        62 &          &         &         &         &         &          &  &          &         &          &       &    &          &    &   &     &             \\ \cmidrule{5-5}
&  & Moderate &     &        47 &          &         &         &         &         &          &  &          &         &          &       &    &          &    &   &     &             \\ \cmidrule{5-5}
&  & Severe &     &        56 &          &         &         &         &         &          &  &          &         &          &       &    &          &    &   &     &             \\
\cmidrule{2-22}
& \multirow{2.5}{*}{PDR} & HR &     &           &      161 &         &         &         &         &          &  &          &         &          &       &    &          &    &   &     &             \\ \cmidrule{6-6}
&  & NHR &     &           &        9 &         &         &         &         &          &  &          &         &          &       &    &          &    &   &     &             \\
\midrule 
\multirow{5}{*}{\rotatebox[origin=l]{90}{\begin{tabular}{@{}c@{}}\textbf{Span}\\\textbf{Validity}
\end{tabular}}} & \multirow{2.5}{*}{ME} & False &     &           &          &         &      12 &         &         &          &  &          &         &          &       &    &          &    &   &     &             \\ \cmidrule{8-8}
&  & True &     &           &          &         &     896 &         &         &          &  &          &         &          &       &    &          &    &   &     &             \\
\cmidrule{2-22}
& \multirow{2.5}{*}{\begin{tabular}{@{}c@{}}Retina\\Surgery\end{tabular}} & False &     &           &          &         &         &         &         &          &  &          &         &          &    37 &    &          &    &   &     &             \\ \cmidrule{16-16}
&  & True &     &           &          &         &         &         &         &          &  &          &         &          &   226 &    &          &    &   &     &             \\
\bottomrule
\end{tabular}
\end{adjustbox}
\caption{Distribution of attribute labels for each of the consolidated tasks, broken down by clinical concept.}
\label{tab:labeldistfull}
\end{table*}

\section{Domain Adaptation Still Matters (\S \ref{sec:exp})} \label{apx:expdesign}

\begin{algorithm}[t]
\small    
\KwIn{$K$ \# of Folds}
\KwData{$g$ group IDs for each document, $Y$ one-hot labels for each document}
\KwOut{$s$ split assignments for each group ID}    
\Begin{
    $g^\prime \leftarrow$ unique group IDs\;\\
    $Y^\prime \leftarrow$ group-level one-hot encoding of labels\;\\
    $s \leftarrow$ list of length $\vert g^\prime \vert$\;\\
    $C[K, \vert\mathcal{L}\vert] \leftarrow$ Count of group IDs with label $\ell \in \mathcal{L}$        assigned to fold $k \in K$\;\\
    \While{not all group IDs assigned in $s$}{
        $m \leftarrow$ unassigned group ID mask\;\\
        $z \leftarrow \sum_{\ell \in \mathcal{L}} Y^\prime[m]$\;\\
        $\ell^* \leftarrow \argmin_{\ell \in \mathcal{L}} z$\;\\
        $p \leftarrow$ sort folds $1 \dots K$ ascending based on $C[k,\ell^*]$\ with ties broken by $C[k,:]$\;\\
        \For{fold $i$ in folds $1 \dots K$}{
            sample $g^* \ni: (g^* \in m) \land (Y^\prime[g^*, \ell^*] = 1)$\;\\
            assign $s[g^*] \leftarrow i$\;\\
            update counts $C$\;
        }
    }
}
\begin{adjustbox}{width=\linewidth}
    \caption{Multi-label, Multi-task Stratification}
    \label{alg:strat}
\end{adjustbox}
\end{algorithm}

\subsection{Stratified Multi-task, Multi-label Cross Validation}

We use a variation of stratified cross-validation (see Algorithm \ref{alg:strat}) for task model experiments, making two modifications to the standard evaluation protocol to minimize information leakage and address concept-attribute class imbalance. First, stratification is done with respect to patients instead of documents (i.e., encounters) or spans. This is done to mitigate the risk of overestimating generalization performance due to copy-forward and other near-duplication that is likely to occur across multiple encounters for the same patient. Pilot experiments confirm that performance estimates are higher without this splitting condition.

Second, assignments to each fold are made in an iterative fashion using a (potentially) different label criteria during each iteration. This is done to account for the multi-task, multi-label nature of the dataset and address concept-attribute imbalance. Unique combinations of target classes are too sparse to use as stratification, while a purely random stratification approach could lead to some folds not having any instances of certain concepts or attribute classes. Although stratification could be done for each classification task independently, this would preclude us from making direct performance comparisons in the event we trained the task models in a multi-task fashion. Our approach is inspired by prior work in multi-label stratification \citep{sechidis2011stratification}. We provide an implementation in the supplemental material.

\subsection{Experimental Setup}

All results in this paper are reported using our stratified cross-validation approach, with $K = 5$ folds. Within each fold, 3 patient subsets are used for training, 1 subset is used for parameter tuning and model selection, and 1 subset is used for evaluation. 

Although the number of unique patients in each subset is roughly equivalent, the number of encounters and concepts spans is not. This is due to the non-uniform concentration of encounters and spans per patient. To limit any single patient from contributing too strongly to the training or evaluation process, we sample a maximum of 10 (Concept, Attribute, Label) tuples from each patient.

\subsection{Compute Environment} \label{apx:model}

All experiments were run in a HIPAA-compliant remote computing environment secured with OS-level group permissions. Language and task models were trained on servers outfitted with NVIDIA Tesla M60 GPUs (2 x 8 GB VRAM) and an Intel Xeon E5-3698 CPU (2.20 GHz base clock).

\subsection{Majority Classifier} \label{apx:model-majority}

Let an input $X$ to one of our task classifiers consist of two mutually exclusive groups -- 1) a text span $T_c$ which indicates the potential mention of a clinical concept $c$ (e.g., NPDR, Retinal Detachment) and 2) a context window $T_z$ of text surrounding $T_c$. $T_z$ can be further decomposed as $T^{\text{(pre)}}_{z}$ and $T^{\text{(post)}}_{z}$, such that input $X$ is the ordered concatenation of the components $X = \langle T^{\text{(pre)}}_{z}, T_c, T^{\text{(post)}}_{z} \rangle$.

A majority classifier $M$ for a classification task with $K$ classes outputs the class $k \in K$ which was seen most frequently during training, regardless of input $X$. In \S \ref{sec:exp}, we consider a variation of a traditional majority classifier, $M^\prime$ which outputs the class $k \in K$ which was seen most frequently during training amongst training instances associated with the same clinical concept $c$ and token span $T_c$ as the input instance $X$.

Due to the small, imbalanced nature of our dataset, we cannot assume that all combinations ($c$, $T_c$) will have been seen at training time. As such, we adopt the logic enumerated below to back-off to a solution depending on the presence of $c$ and $T_c$ in our training data. There are four cases we must consider.

\begin{enumerate}
    \item ($c$ seen, $T_c$ seen): $M^\prime$ outputs the most common class label $k$ seen amongst all training inputs $X$ having $c$ and $T_c$.
    \item ($c$ seen, $T_c$ not seen): $M^\prime$ outputs the most common class label $k$ seen amongst all training inputs $X$ having $c$.
    \item ($c$ not seen, $T_c$ seen): Our expressions are set up such that $T_c \Rightarrow c$. Therefore, this case does not occur in practice. However, $M^\prime$ would output the most common class label $k$ seen amongst all training inputs $X$ having $T_{c}$.
    \item ($c$ not seen, $T_c$ not seen): $M^\prime$ outputs the most common class label $k$ seen amongst all training inputs $X$, regardless of $c$ or $T_c$. Accordingly, $M^\prime \equiv M$.
\end{enumerate}

The primary purpose of the majority classifier is to provide an appropriate reference for performance given the imbalanced nature of many tasks in the study. That said, the majority model $M^\prime$ conditioned on $c$ and $T_c$ will outperform a simple majority model $M$ without additional conditioning when the extracted text spans can be directly associated with a downstream attribute (e.g., T2DM = Type 2 Diabetes Mellitus, NVI = Neovascularization of the Iris, Trabeculectomy = NVG Surgery Type) or when there are different concept-level differences in the class distribution within a task.

\subsection{Language Models} \label{apx:model-lm}

We use Hugging Face's implementation of BERT via the \texttt{transformers} Python package. Base BERT \citep{devlin2018bert} and Clinical BERT \citep{alsentzer2019publicly} are initialized from the Hugging Face hub. Although we use a context window of 128-tokens, we do not re-initialize the positional embeddings.

\begin{figure}[t]
    \centering
    \includegraphics[width=\linewidth]{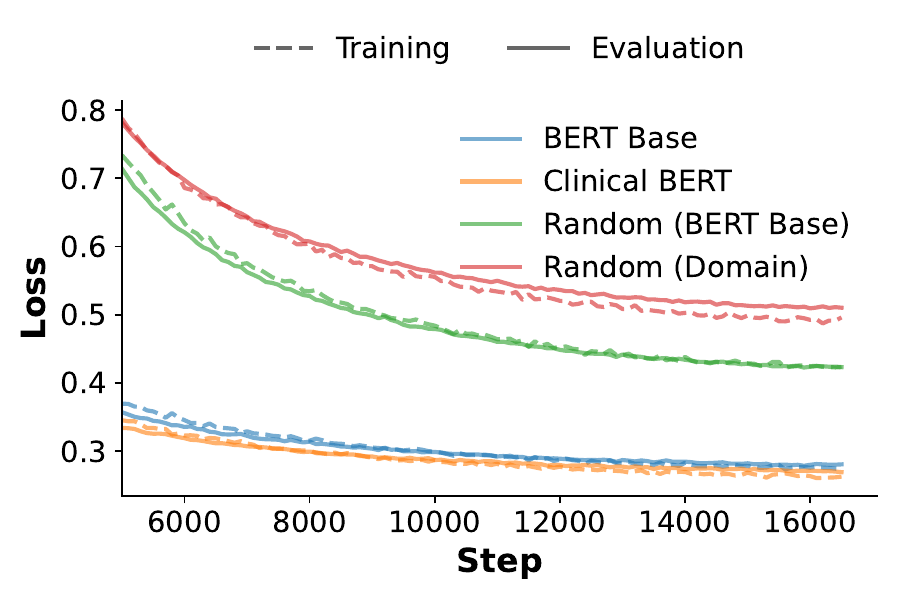}
    \caption{Training and validation loss curves for continued pretraining on the full dataset as a function of initialization and tokenizer. We start the x-axis at step 5,000 for visual clarity.}
    \label{fig:lmloss-init}
\end{figure}

\begin{figure}[t]
    \centering
    \includegraphics[width=\linewidth]{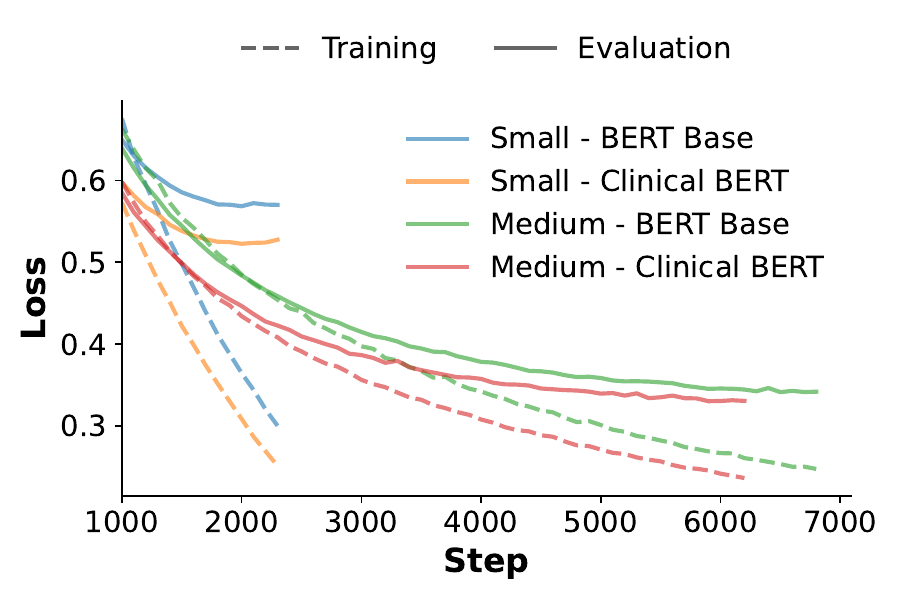}
    \caption{Training and validation loss curves for continued pretraining on the smaller subsets of the full dataset. We start the x-axis at step 1,000 for visual clarity.}
    \label{fig:lmloss-size}
\end{figure}

All patients not included in the annotated dataset are considered candidates for continued pretraining. We use notes from a randomly sampled 5\% of this cohort as an evaluation set to monitor convergence, training on notes from the remaining 95\% of patients. We continue pretraining of BERT Base and Clinical BERT for a maximum of 16,500 steps using the AdamW optimizer \citep{loshchilov2017decoupled}, an initial learning rate of 5e-5, linear learning rate decay with 5,000 warmup steps, weight decay of 0.01, and an effective batch size of 1,024 (via distributed data parallelism and gradient accumulation). We also apply early stopping based on evaluation loss with a tolerance of 0.01 and patience of 3 (evaluated every 100 steps). The original tokenizer and vocabulary of two BERT models is not modified. We use the masked language modeling objective with a masking probability of 0.15 (up to maximally 20 tokens / sequence) and binary cross entropy loss. All sequences are a maximum 128 tokens long. All other parameters in the \texttt{Trainer} and \texttt{TrainingArguments} modules of the \texttt{transformers} package are kept at their defaults. Loss curves as a function of the initialization weights and tokenizer are provided in Figure \ref{fig:lmloss-init}. Loss curves for BERT Base and Clinical BERT pretrained on subsets of the full dataset are provided in Figure \ref{fig:lmloss-size}.

\subsection{Task Models} \label{apx:model-task}

\begin{figure}[t]
    \centering
    \includegraphics[width=\linewidth]{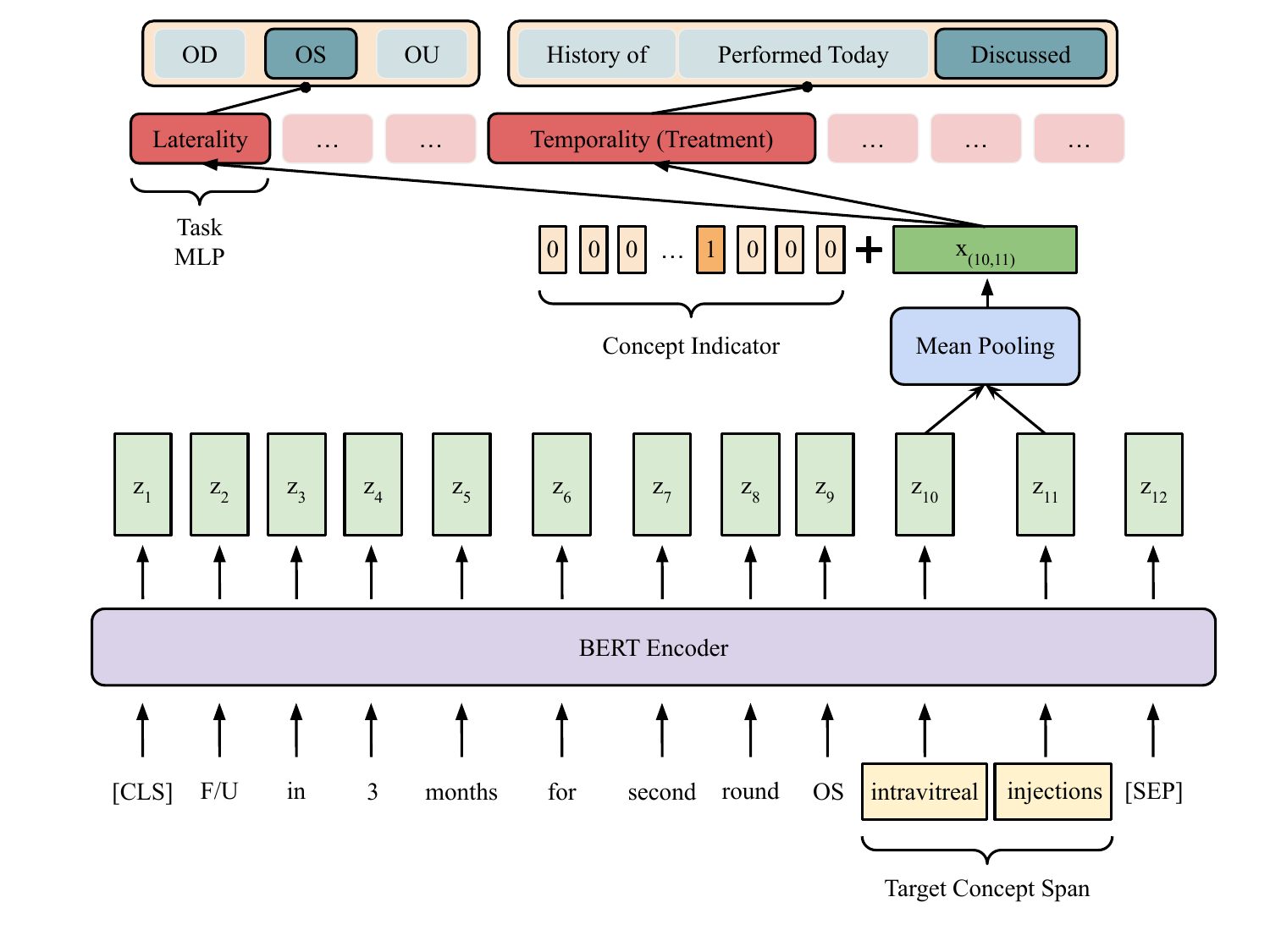}
    \caption{Overview of our model architecture. In practice, we center the context window around each target concept span and train each task model independently.}
    \label{fig:architecture}
\end{figure}

The reader should recall two important aspects of the annotated data and concept ontology. First, there may be multiple instances of a clinical concept in a single encounter, not all of which share the same attributes (i.e., laterality, severity, temporality). Second, regular expressions may identify overlapping text spans for \emph{different} clinical concepts. These overlaps may be partial -- e.g., [[diabetic] retinopathy] -- or full  -- e.g., [E11.319] is a match for both diabetic retinopathy and diabetes mellitus. Together, these aspects inspire us to treat the task of inferring attributes associated with extracted clinical concepts as a span classification problem as opposed to either a document-classification or token-classification problem.

We provide an overview of our model architecture in Figure \ref{fig:architecture}. A maximum of 128 tokens centered around the target concept span are passed through the BERT encoder. Centering is accomplished by iteratively expanding the context window on the left and right of the target concept span until the maximum context size or a boundary of the note has been reached. Embeddings for tokens in the target concept span are mean-pooled and then concatenated with a one-hot indicator vector denoting the clinical concept being classified. The 19-dimensional concept-indicator (one dimension per clinical concept) is included to account for cases of fully-overlapping concept spans for multiple target concepts and explicitly model concept-specific class priors. Finally, the concatenated vectors are passed through a dense, one-layer MLP. The MLP has an input dimensionality of 787 (768 dimension BERT output layer $+$ 19 dimension concept-indicator) and a hidden dimension of either 0 (i.e., linear map to output), 256, or 512 (see discussion below regarding hyperparameter tuning).

All task models are trained independently. That is, a separate backbone LM encoder and MLP classification head are trained for each task. Multi-task learning (i.e., training a shared LM encoder with independent MLP classification heads for each task) is out of this project's scope. We opted to focus on single-task learning because multi-task learning can introduce additional optimization challenges and result in negative task transfer \citep{lee2016asymmetric,wu2020understanding}. That said, future work which explores the interaction between pretraining domain and the number of fine-tuning tasks is warranted.

Both training and evaluation are implemented in \texttt{torch} and leverage language models from the \texttt{transformers} Python package. We use the same training setup for all tasks. We minimize the cross entropy loss, inversely weighted based on training class proportions \citep{king2001logistic}. We use the AdamW optimizer and a step-wise learning rate scheduler configured to reduce the learning rate 10\% every 50 steps after a warmup of 100 steps. We hold the weight decay factor (0.1), dropout rate (0.1), and gradient clipping max norm (1.0) constant.

Models are trained for a minimum of 50 steps and a maximum of 500 steps, with early stopping configured to preempt training if the validation loss or macro F1-score have not improved by 1\% over 5 evaluation subroutines. We evaluate performance every 5 steps and select the checkpoint which maximizes macro F1-score in the validation set. 

We run a hyperparameter grid search over the Cartesian product of learning rates \{1e-5, 1e-4, 1e-3\}, batch sizes \{8, 32, 64\}, and MLP hidden dimensions \{Null, 256, 512\}.\footnote{A null hidden dimension denotes a single linear layer from the pooled embeddings to the output layer.} The search is run independently for each cross validation fold. The hyperparameter configuration which maximizes macro F1-score in a fold's associated development set is used for evaluation on the test set. Configurations which the training data does not support (i.e., a batch size larger than the number of training instances) are ignored. Higher learning rates were better for task models with frozen BERT encoders, while lower learning rates were better for task models with unfrozen BERT encoders. No other clear trends where observed.

\subsection{Task-specific Outcomes} \label{apx:performance-task-specific}

We provide task-specific performance breakdowns for the pretraining dataset size (\S \ref{sec:exp-size}) and weight initialization (\S \ref{sec:exp-scratch}) experiments in Tables \ref{tab:performance-size} and \ref{tab:performance-initialization}, respectively. Fold-level performance measures and statistical test results for all tasks are included in our digital supplement.\footnotemark[9]

\begin{table*}[t]
    \centering
    \begin{adjustbox}{width=\linewidth}
    \begin{tabular}{c c c c c c c c c c} \toprule
    &
        &
        \multicolumn{8}{c}{\textbf{Frozen Encoder} (\raisebox{-0.5ex}{\SnowflakeChevron})} \\ \cmidrule(lr){3-10}
    &
        &
        \multicolumn{2}{c}{\texttt{Zero}}  &
        \multicolumn{2}{c}{\texttt{Small}} &
        \multicolumn{2}{c}{\texttt{Medium}} &
        \multicolumn{2}{c}{\texttt{Large}} \\ \cmidrule(lr){3-4} \cmidrule(lr){5-6}  \cmidrule(lr){7-8} \cmidrule(lr){9-10}
    \textbf{Attribute} &
            \textbf{Concepts} &
            \textbf{Base} &
            \textbf{Clinical} &
            \textbf{Base} &
            \textbf{Clinical} &
            \textbf{Base} &
            \textbf{Clinical} &
            \textbf{Base} &
            \textbf{Clinical} \\ \midrule
    \multirow{3}{*}{Temporality} &
        Retina & 
        .81 \textsubscript{(.79,.82)} &
        .83 \textsubscript{(.82,.84)} &
        .82 \textsubscript{(.80,.83)} &
        .85 \textsubscript{(.84,.87)} &
        .84 \textsubscript{(.82,.85)} &
        .85 \textsubscript{(.84,.87)} &
        .84 \textsubscript{(.83,.86)} &
        .85 \textsubscript{(.84,.87)}\\
    &
        DM Complications &
        .81 \textsubscript{(.73,.88)} &
        .80 \textsubscript{(.70,.89)} &
        .87 \textsubscript{(.84,.91)} &
        .86 \textsubscript{(.83,.88)} &
        .82 \textsubscript{(.71,.89)} &
        .82 \textsubscript{(.70,.93)} &
        .81 \textsubscript{(.71,.88)} &
        .84 \textsubscript{(.73,.93)} \\
    & 
        Treatment &
        .59 \textsubscript{(.55,.64)} &
        .69 \textsubscript{(.65,.73)} &
        .64 \textsubscript{(.57,.71)} &
        .76 \textsubscript{(.71,.81)} &
        .78 \textsubscript{(.73,.82)} &
        .77 \textsubscript{(.74,.79)} &
        .81 \textsubscript{(.79,.83)} &
        .81 \textsubscript{(.77,.83)} \\ \cmidrule{1-10}
    \multirow{1}{*}{Laterality} &
        All &
        .54 \textsubscript{(.51,.57)} &
        .56 \textsubscript{(.54,.58)} &
        .56 \textsubscript{(.54,.58)} &
        .56 \textsubscript{(.53,.60)} &
        .59 \textsubscript{(.56,.62)} &
        .59 \textsubscript{(.57,.62)} &
        .60 \textsubscript{(.58,.61)} &
        .60 \textsubscript{(.57,.63)} \\ \cmidrule{1-10}
    \multirow{6}{*}{Type} &
        ME &
        .83 \textsubscript{(.79,.88)} &
        .82 \textsubscript{(.78,.86)} &
        .83 \textsubscript{(.80,.86)} &
        .81 \textsubscript{(.75,.86)} &
        .88 \textsubscript{(.84,.92)} &
        .82 \textsubscript{(.79,.86)} &
        .86 \textsubscript{(.79,.91)} &
        .85 \textsubscript{(.82,.89)} \\
    &
        RD &
        .75 \textsubscript{(.57,.89)} &
        .78 \textsubscript{(.59,.94)} &
        .79 \textsubscript{(.66,.90)} &
        .79 \textsubscript{(.60,.94)} &
        .77 \textsubscript{(.64,.88)} &
        .72 \textsubscript{(.49,.94)} &
        .70 \textsubscript{(.55,.83)} &
        .72 \textsubscript{(.50,.95)} \\
    &
        NV &
        .66 \textsubscript{(.53,.80)} &
        .71 \textsubscript{(.56,.87)} &
        .79 \textsubscript{(.71,.89)} &
        .80 \textsubscript{(.66,.93)} &
        .82 \textsubscript{(.73,.91)} &
        .85 \textsubscript{(.78,.92)} &
        .81 \textsubscript{(.71,.91)} &
        .81 \textsubscript{(.73,.90)} \\
    & 
        DM &
        .39 \textsubscript{(.31,.53)} &
        .31 \textsubscript{(.28,.35)} &
        .31 \textsubscript{(.30,.32)} &
        .38 \textsubscript{(.29,.52)} &
        .39 \textsubscript{(.31,.53)} &
        .32 \textsubscript{(.28,.37)} &
        .37 \textsubscript{(.29,.52)} &
        .33 \textsubscript{(.29,.38)} \\
    &
        NVG Surgery &
        .85 \textsubscript{(.69,1.0)} &
        .79 \textsubscript{(.54,1.0)} &
        .82 \textsubscript{(.62,1.0)} &
        .79 \textsubscript{(.61,.96)} &
        .71 \textsubscript{(.51,.93)} &
        .79 \textsubscript{(.51,1.0)} &
        .74 \textsubscript{(.63,.88)} &
        .79 \textsubscript{(.63,.96)} \\
    &
        Retina Surgery &
        .51 \textsubscript{(.38,.64)} &
        .59 \textsubscript{(.46,.70)} &
        .63 \textsubscript{(.45,.78)} &
        .52 \textsubscript{(.40,.65)} &
        .50 \textsubscript{(.26,.67)} &
        .52 \textsubscript{(.37,.66)} &
        .60 \textsubscript{(.44,.71)} &
        .64 \textsubscript{(.53,.75)} \\ \cmidrule{1-10}
    \multirow{2}{*}{Severity} &
        NPDR &
        .56 \textsubscript{(.50,.62)} &
        .58 \textsubscript{(.54,.61)} &
        .55 \textsubscript{(.51,.61)} &
        .60 \textsubscript{(.55,.65)} &
        .64 \textsubscript{(.53,.79)} &
        .74 \textsubscript{(.64,.84)} &
        .69 \textsubscript{(.57,.84)} &
        .70 \textsubscript{(.60,.83)} \\
    &
        PDR &
        .45 \textsubscript{(.43,.48)} &
        .43 \textsubscript{(.37,.47)} &
        .38 \textsubscript{(.32,.43)} &
        .45 \textsubscript{(.42,.47)} &
        .44 \textsubscript{(.42,.46)} &
        .53 \textsubscript{(.44,.68)} &
        .39 \textsubscript{(.29,.47)} &
        .45 \textsubscript{(.43,.48)} \\ \cmidrule{1-10}
    \multirow{2}{*}{Span Validity} &
        ME &
        .60 \textsubscript{(.49,.72)} &
        .65 \textsubscript{(.52,.78)} &
        .56 \textsubscript{(.49,.64)} &
        .74 \textsubscript{(.56,.91)} &
        .71 \textsubscript{(.55,.87)} &
        .76 \textsubscript{(.56,.97)} &
        .77 \textsubscript{(.57,.97)} &
        .66 \textsubscript{(.49,.86)} \\
    &    
        Retina Surgery &
        .77 \textsubscript{(.72,.81)} &
        .76 \textsubscript{(.69,.84)} &
        .81 \textsubscript{(.72,.89)} &
        .77 \textsubscript{(.69,.84)} &
        .80 \textsubscript{(.73,.88)} &
        .84 \textsubscript{(.78,.89)} &
        .83 \textsubscript{(.77,.89)} &
        .83 \textsubscript{(.76,.90)} \\ \cmidrule{1-10}
    \multicolumn{2}{c}{\textbf{Average (All Tasks)}} &
        .65 \textsubscript{(.61,.69)} &
        .67 \textsubscript{(.62,.71)} &
        .67 \textsubscript{(.62,.71)} &
        .69 \textsubscript{(.65,.73)} &
        .69 \textsubscript{(.64,.74)} &
        .71 \textsubscript{(.66,.75)} &
        .70 \textsubscript{(.65,.75)} &
        .71 \textsubscript{(.66,.75)} \\  \bottomrule
    \multicolumn{10}{c}{} \\ \toprule
    &
        &
        \multicolumn{8}{c}{\textbf{Unfrozen Encoder} (\raisebox{-0.5ex}{\FilledRainCloud})} \\ \cmidrule(lr){3-10}
    &
        &
        \multicolumn{2}{c}{\texttt{Zero}}  &
        \multicolumn{2}{c}{\texttt{Small}} &
        \multicolumn{2}{c}{\texttt{Medium}} &
        \multicolumn{2}{c}{\texttt{Large}} \\ \cmidrule(lr){3-4} \cmidrule(lr){5-6}  \cmidrule(lr){7-8} \cmidrule(lr){9-10}
    \textbf{Attribute} &
            \textbf{Concepts} &
            \textbf{Base} &
            \textbf{Clinical} &
            \textbf{Base} &
            \textbf{Clinical} &
            \textbf{Base} &
            \textbf{Clinical} &
            \textbf{Base} &
            \textbf{Clinical} \\ \midrule
    \multirow{3}{*}{Temporality} &
        Retina & 
        .83 \textsubscript{(.82,.84)} &
        .84 \textsubscript{(.83,.86)} &
        .85 \textsubscript{(.83,.87)} &
        .85 \textsubscript{(.84,.86)} &
        .87 \textsubscript{(.85,.88)} &
        .87 \textsubscript{(.85,.88)} &
        .87 \textsubscript{(.85,.89)} &
        .87 \textsubscript{(.85,.88)} \\
    &
        DM Complications &
        .80 \textsubscript{(.70,.89)} &
        .84 \textsubscript{(.76,.90)} &
        .79 \textsubscript{(.70,.87)} &
        .86 \textsubscript{(.82,.90)} &
        .89 \textsubscript{(.85,.93)} &
        .88 \textsubscript{(.84,.92)} &
        .80 \textsubscript{(.71,.88)} &
        .85 \textsubscript{(.77,.92)} \\
    & 
        Treatment &
        .79 \textsubscript{(.75,.82)} &
        .81 \textsubscript{(.78,.84)} &
        .80 \textsubscript{(.76,.84)} &
        .84 \textsubscript{(.81,.86)} &
        .83 \textsubscript{(.79,.86)} &
        .85 \textsubscript{(.81,.87)} &
        .84 \textsubscript{(.81,.86)} &
        .82 \textsubscript{(.76,.85)} \\ \cmidrule{1-10}
    \multirow{1}{*}{Laterality} &
        All &
        .84 \textsubscript{(.83,.86)} &
        .84 \textsubscript{(.81,.87)} &
        .87 \textsubscript{(.86,.87)} &
        .87 \textsubscript{(.85,.89)} &
        .90 \textsubscript{(.89,.92)} &
        .89 \textsubscript{(.89,.90)} &
        .92 \textsubscript{(.90,.93)} &
        .90 \textsubscript{(.89,.92)} \\ \cmidrule{1-10}
    \multirow{6}{*}{Type} &
        ME &
        .87 \textsubscript{(.79,.93)} &
        .87 \textsubscript{(.80,.93)} &
        .86 \textsubscript{(.80,.93)} &
        .89 \textsubscript{(.86,.92)} &
        .90 \textsubscript{(.84,.95)} &
        .91 \textsubscript{(.89,.94)} &
        .88 \textsubscript{(.82,.94)} &
        .90 \textsubscript{(.85,.94)} \\
    &
        RD &
        .79 \textsubscript{(.58,.95)} &
        .81 \textsubscript{(.61,.98)} &
        .79 \textsubscript{(.66,.91)} &
        .83 \textsubscript{(.67,.95)} &
        .80 \textsubscript{(.69,.90)} &
        .81 \textsubscript{(.61,.99)} &
        .82 \textsubscript{(.76,.88)} &
        .87 \textsubscript{(.78,.95)} \\
    &
        NV &
        .75 \textsubscript{(.60,.89)} &
        .78 \textsubscript{(.65,.92)} &
        .77 \textsubscript{(.62,.89)} &
        .76 \textsubscript{(.63,.89)} &
        .78 \textsubscript{(.64,.92)} &
        .78 \textsubscript{(.66,.89)} &
        .82 \textsubscript{(.72,.93)} &
        .77 \textsubscript{(.66,.86)} \\
    & 
        DM &
        .40 \textsubscript{(.32,.52)} &
        .40 \textsubscript{(.31,.53)} &
        .35 \textsubscript{(.32,.38)} &
        .40 \textsubscript{(.31,.53)} &
        .59 \textsubscript{(.48,.77)} &
        .52 \textsubscript{(.37,.76)} &
        .57 \textsubscript{(.43,.79)} &
        .54 \textsubscript{(.40,.74)} \\
    &
        NVG Surgery &
        .85 \textsubscript{(.69,1.0)} &
        .85 \textsubscript{(.69,1.0)} &
        .85 \textsubscript{(.69,1.0)} &
        .85 \textsubscript{(.69,1.0)} &
        .85 \textsubscript{(.69,1.0)} &
        .85 \textsubscript{(.69,1.0)} &
        .85 \textsubscript{(.69,1.0)} &
        .85 \textsubscript{(.69,1.0)} \\
    &
        Retina Surgery &
        .66 \textsubscript{(.52,.79)} &
        .52 \textsubscript{(.46,.59)} &
        .62 \textsubscript{(.50,.73)} &
        .61 \textsubscript{(.53,.68)} &
        .70 \textsubscript{(.58,.79)} &
        .70 \textsubscript{(.63,.75)} &
        .71 \textsubscript{(.59,.83)} &
        .76 \textsubscript{(.65,.85)} \\ \cmidrule{1-10}
    \multirow{2}{*}{Severity} &
        NPDR &
        .83 \textsubscript{(.73,.90)} &
        .91 \textsubscript{(.87,.96)} &
        .92 \textsubscript{(.86,.98)} &
        .91 \textsubscript{(.85,.96)} &
        .91 \textsubscript{(.86,.97)} &
        .96 \textsubscript{(.90,1.0)} &
        .89 \textsubscript{(.77,.98)} &
        .95 \textsubscript{(.90,.99)} \\
    &
        PDR &
        .71 \textsubscript{(.53,.89)} &
        .53 \textsubscript{(.36,.77)} &
        .67 \textsubscript{(.45,.90)} &
        .69 \textsubscript{(.51,.86)} &
        .73 \textsubscript{(.55,.90)} &
        .69 \textsubscript{(.47,.90)} &
        .81 \textsubscript{(.64,.93)} &
        .82 \textsubscript{(.63,.98)} \\ \cmidrule{1-10}
    \multirow{2}{*}{Span Validity} &
        ME &
        .55 \textsubscript{(.49,.62)} &
        .56 \textsubscript{(.50,.64)} &
        .57 \textsubscript{(.49,.73)} &
        .63 \textsubscript{(.50,.78)} &
        .73 \textsubscript{(.60,.88)} &
        .82 \textsubscript{(.65,.97)} &
        .81 \textsubscript{(.64,.97)} &
        .83 \textsubscript{(.65,.98)} \\
    &    
        Retina Surgery &
        .77 \textsubscript{(.68,.87)} &
        .80 \textsubscript{(.74,.86)} &
        .75 \textsubscript{(.68,.83)} &
        .81 \textsubscript{(.76,.86)} &
        .76 \textsubscript{(.68,.83)} &
        .87 \textsubscript{(.82,.90)} &
        .82 \textsubscript{(.75,.90)} &
        .80 \textsubscript{(.75,.88)} \\ \cmidrule{1-10}
    \multicolumn{2}{c}{\textbf{Average (All Tasks)}} &
        .75 \textsubscript{(.70,.79)} &
        .74 \textsubscript{(.69,.79)} &
        .75 \textsubscript{(.70,.79)} &
        .77 \textsubscript{(.73,.81)} &
        .80 \textsubscript{(.76,.84)} &
        .81 \textsubscript{(.77,.85)} &
        .82 \textsubscript{(.78,.85)} &
        .82 \textsubscript{(.79,.86)} \\ \bottomrule
    \end{tabular}
    \end{adjustbox}
    \caption{Task-specific performance (macro F1) as a function of the pretraining dataset size. Performance with a frozen encoder and an unfrozen encoder is shown in the top and bottom tables, respectively. We observe gradual increases in performance for both BERT Base and Clinical BERT task models as the pretraining dataset grows. The Clinical BERT model is able to take advantage of the \texttt{Small} pretraining dataset slightly better than BERT Base.}
    \label{tab:performance-size}
\end{table*}

\begin{table*}[t]
    \centering
    \begin{adjustbox}{width=\linewidth}
    \small
    \begin{tabular}{c c c c c c c c} \toprule
    &
        &
        \multicolumn{3}{c}{\textbf{Frozen Encoder} (\raisebox{-0.5ex}{\SnowflakeChevron})} &
        \multicolumn{3}{c}{\textbf{Unfrozen Encoder} (\raisebox{-0.5ex}{\FilledRainCloud})} \\ \cmidrule(lr){3-5} \cmidrule(lr){6-8}
    &
        &
        \multicolumn{1}{c}{\textbf{BERT Base}} &
        \multicolumn{2}{c}{\textbf{Random}} &
        \multicolumn{1}{c}{\textbf{BERT Base}} &
        \multicolumn{2}{c}{\textbf{Random}} \\ \cmidrule(lr){3-3} \cmidrule(lr){4-5} \cmidrule(lr){6-6} \cmidrule(lr){7-8}
    \textbf{Concept} &
        \textbf{Attribute} &
        \multicolumn{1}{c}{\textbf{BERT Base}} &
        \multicolumn{1}{c}{\textbf{BERT Base}} &
        \multicolumn{1}{c}{\textbf{Learned}} &
        \multicolumn{1}{c}{\textbf{BERT Base}} &
        \multicolumn{1}{c}{\textbf{BERT Base}} &
        \multicolumn{1}{c}{\textbf{Learned}} \\ \midrule
    \multirow{3}{*}{Temporality} &
        Retina & 
        .84 \textsubscript{(.83,.86)} &
        .83 \textsubscript{(.82,.84)} &
        .86 \textsubscript{(.84,.87)} &
        .87 \textsubscript{(.85,.89)} &
        .85 \textsubscript{(.82,.87)} &
        .85 \textsubscript{(.83,.87)} \\
    &
        DM Complications &
        .81 \textsubscript{(.71,.88)} &
        .79 \textsubscript{(.65,.89)} &
        .80 \textsubscript{(.75,.84)} &
        .80 \textsubscript{(.71,.88)} &
        .83 \textsubscript{(.73,.92)} &
        .90 \textsubscript{(.87,.93)} \\
    & 
        Treatment &
        .81 \textsubscript{(.79,.83)} &
        .72 \textsubscript{(.70,.74)} &
        .80 \textsubscript{(.77,.83)} &
        .84 \textsubscript{(.81,.86)} &
        .80 \textsubscript{(.77,.84)} &
        .79 \textsubscript{(.72,.85)} \\ \cmidrule{1-8}
    \multirow{1}{*}{Laterality} &
        All &
        .60 \textsubscript{(.58,.61)} &
        .58 \textsubscript{(.56,.60)} &
        .61 \textsubscript{(.60,.62)} &
        .92 \textsubscript{(.90,.93)} &
        .87 \textsubscript{(.85,.89)} &
        .89 \textsubscript{(.87,.91)} \\ \cmidrule{1-8}
    \multirow{6}{*}{Type} &
        ME &
        .86 \textsubscript{(.79,.91)} &
        .87 \textsubscript{(.82,.91)} &
        .89 \textsubscript{(.85,.93)} &
        .88 \textsubscript{(.82,.94)} &
        .89 \textsubscript{(.85,.92)} &
        .90 \textsubscript{(.87,.94)} \\
    &
        RD &
        .70 \textsubscript{(.55,.83)} &
        .80 \textsubscript{(.60,.97)} &
        .84 \textsubscript{(.77,.91)} &
        .82 \textsubscript{(.76,.88)} &
        .76 \textsubscript{(.58,.92)} &
        .74 \textsubscript{(.65,.83)} \\
    &
        NV &
        .81 \textsubscript{(.71,.91)} &
        .83 \textsubscript{(.74,.93)} &
        .81 \textsubscript{(.74,.89)} &
        .82 \textsubscript{(.72,.93)} &
        .79 \textsubscript{(.71,.87)} &
        .75 \textsubscript{(.69,.83)} \\
    & 
        DM &
        .37 \textsubscript{(.29,.52)} &
        .42 \textsubscript{(.32,.58)} &
        .40 \textsubscript{(.31,.54)} &
        .57 \textsubscript{(.43,.79)} &
        .40 \textsubscript{(.29,.54)} &
        .53 \textsubscript{(.38,.75)} \\
    &
        NVG Surgery &
        .74 \textsubscript{(.63,.88)} &
        .72 \textsubscript{(.43,.97)} &
        .75 \textsubscript{(.58,.93)} &
        .85 \textsubscript{(.69,1.0)} &
        .85 \textsubscript{(.69,1.0)} &
        .85 \textsubscript{(.69,1.0)} \\
    &
        Retina Surgery &
        .60 \textsubscript{(.44,.71)} &
        .67 \textsubscript{(.53,.80)} &
        .53 \textsubscript{(.42,.63)} &
        .71 \textsubscript{(.59,.83)} &
        .71 \textsubscript{(.63,.79)} &
        .75 \textsubscript{(.63,.87)} \\ \cmidrule{1-8}
    \multirow{2}{*}{Severity} &
        NPDR &
        .69 \textsubscript{(.57,.84)} &
        .73 \textsubscript{(.65,.84)} &
        .67 \textsubscript{(.46,.86)} &
        .89 \textsubscript{(.77,.98)} &
        .84 \textsubscript{(.73,.91)} &
        .91 \textsubscript{(.88,.94)} \\
    &
        PDR &
        .39 \textsubscript{(.29,.47)} &
        .43 \textsubscript{(.35,.48)} &
        .48 \textsubscript{(.47,.48)} &
        .81 \textsubscript{(.64,.93)} &
        .66 \textsubscript{(.43,.90)} &
        .89 \textsubscript{(.79,.98)} \\ \cmidrule{1-8}
    \multirow{2}{*}{Span Validity} &
        ME &
        .77 \textsubscript{(.57,.97)} &
        .74 \textsubscript{(.54,.94)} &
        .72 \textsubscript{(.53,.90)} &
        .81 \textsubscript{(.64,.97)} &
        .75 \textsubscript{(.60,.90)} &
        .82 \textsubscript{(.66,.95)} \\
    &    
        Retina Surgery &
        .83 \textsubscript{(.77,.89)} &
        .77 \textsubscript{(.59,.89)} &
        .83 \textsubscript{(.77,.90)} &
        .82 \textsubscript{(.75,.90)} &
        .84 \textsubscript{(.81,.88)} &
        .81 \textsubscript{(.74,.87)} \\ \cmidrule{1-8}
    \multicolumn{2}{c}{\textbf{Average (All Tasks)}} &
        .70 \textsubscript{(.65,.75)} &
        .71 \textsubscript{(.66,.75)} &
        .71 \textsubscript{(.67,.76)} &
        .82 \textsubscript{(.78,.85)} &
        .77 \textsubscript{(.73,.81)} &
        .81 \textsubscript{(.78,.84)} \\ \bottomrule
    \end{tabular}
    \end{adjustbox}
    \caption{A comparison of task-specific performance (macro F1) when pretraining from scratch instead of continuing pretraining from an existing checkpoint. With the domain-specific (learned) vocabulary, we are able to achieve the same level of performance when pretraining from scratch as we do when pretraining from the existing BERT Base checkpoint.}
    \label{tab:performance-initialization}
\end{table*}

\end{document}